\documentclass{article}
\PassOptionsToPackage{numbers, compress}{natbib}
\usepackage[main, final]{neurips_2025}

\usepackage[utf8]{inputenc} 
\usepackage[T1]{fontenc}    
\usepackage{hyperref}       
\usepackage{url}            
\usepackage{booktabs}       
\usepackage{amsfonts}       
\usepackage{nicefrac}       
\usepackage{microtype}      
\usepackage{xcolor}         
\usepackage{multirow}
\usepackage{rotating} 
\usepackage{listings}
\usepackage{array, cellspace}
\usepackage{subfigure} 
\usepackage{makecell}
\usepackage{longtable}
\usepackage{float}
\usepackage{tabularx}

\newcommand{\re}[1]{{\color{black}#1}}
\newcommand{\nameSeq}{TPTU-SA}
\newcommand{\nameOnestep}{TPTU-OA}

\title{TPTU: Large Language Model-based AI Agents for Task Planning and Tool Usage}

\author{%
  Jingqing Ruan$^\dagger$$^\ddagger$ \\[0.3em]
  ruanjingqing2019@ia.ac.cn
  \And
  Yihong Chen$^\dagger$$^\ddagger$ \\
  chenyihong@sensetime.com
  \And
  Bin Zhang$^\dagger$$^\ddagger$ \\
  zhangbin11@sensetime.com
  \And
  Zhiwei Xu$^\dagger$$^\ddagger$ \\
  xuzhiwei@sensetime.com
  \And
  Tianpeng Bao$^\dagger$ \\
  baotianpeng@sensetime.com
  \And
  Guoqing Du$^\dagger$ \\
  duguoqing@sensetime.com
  \And
  Shiwei Shi$^\dagger$ \\
  shishiwei@sensetime.com
  \And
  Hangyu Mao$^\dagger$$^*$ \\
  maohangyu@sensetime.com
  \And
  Ziyue Li $^+$ \\
  zlibn@connect.ust.hk
  \And
  Xingyu Zeng \\
  zengxingyu@sensetime.com
  \And
  Rui Zhao \\
  zhaorui@sensetime.com
  \\[1em]
  University of Chinese Academy of Sciences, China \\
  SenseTime Research, HKUST, Hong Kong
}

\begin{document}
\maketitle

\def\thefootnote{$^\dagger$}\footnotetext{These authors contribute equally to this work.}\def\thefootnote{\arabic{footnote}}
\def\thefootnote{$^+$}\footnotetext{External discussion and ideation.}\def\thefootnote{\arabic{footnote}}
\def\thefootnote{$^\ddagger$}\footnotetext{These authors work as research interns at SenseTime Research.}\def\thefootnote{\arabic{footnote}}
\def\thefootnote{$^*$}\footnotetext{The corresponding author.}\def\thefootnote{\arabic{footnote}}

\begin{abstract}
With recent advancements in natural language processing, Large Language Models (LLMs) have emerged as powerful tools for various real-world applications. Despite their powers, the intrinsic generative abilities of LLMs may prove insufficient for handling complex tasks, which necessitate a combination of task planning and the usage of external tools. In this paper, we first propose a structured framework tailored for LLM-based AI Agents and then discuss the crucial capabilities necessary for tackling intricate problems. Within this framework, we design two distinct types of agents (i.e., one-step agent and sequential agent) to execute the inference process. Subsequently, we instantiate the framework using various LLMs and evaluate their \underline{T}ask \underline{P}lanning and \underline{T}ool \underline{U}sage (TPTU) abilities on typical tasks. By highlighting key findings and challenges, our goal is to provide a helpful resource for researchers and practitioners to leverage the power of LLMs in their AI applications. Our study emphasizes the substantial potential of these models while also identifying areas that need more investigation and improvement. {The code and resources will be available on GitHub.}
\end{abstract}

\section{Introduction}
Large Language Model (LLM) \cite{zhao2023survey} is a recent breakthrough in natural language processing (NLP) research. These models are trained on massive amounts of text data and can solve a wide range of tasks, even those that were not included in their training dataset, \re{known as ``emerging'' ability}. This ability is especially evident in \re{the tasks of} few-shot \cite{brown2020language} and zero-shot \cite{wei2021finetuned} learning, where LLMs can perform well with minimal or even no \re{fine-tuning to adapt to a new task}. 

\begin{figure}[t]
\centering
\includegraphics[width=0.5\columnwidth]{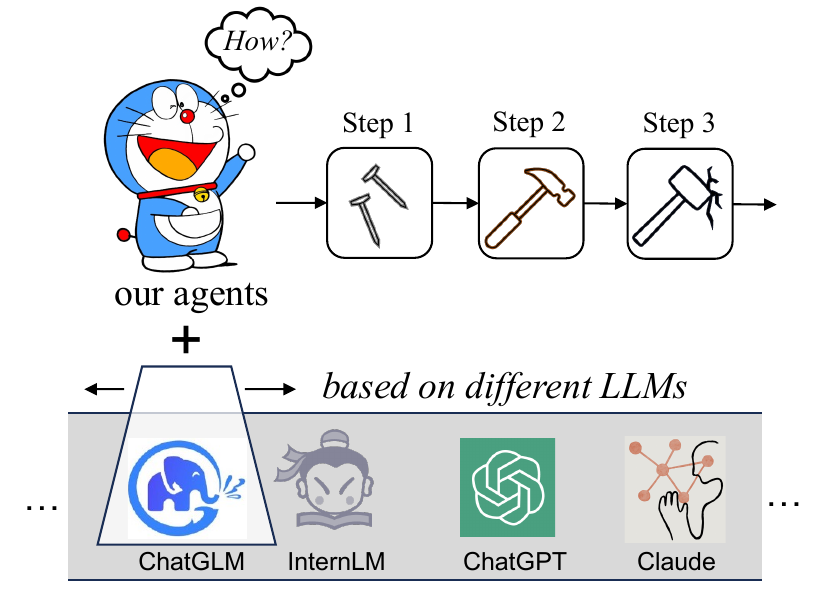}
\caption{Our LLM-based agents plan tasks and use tools.}
\label{fig:intro}
\end{figure}

However, the application of LLMs in real-world settings presents unique challenges. On the one hand, LLMs have proved to be \re{incompetent in} solving logic problems such as mathematics, and their training data is also out of date \re{(e.g., the knowledge cutoff date for GPT-4 \cite{openai2023gpt4} is up to January 2022)}. Teaching LLMs to use tools such as calculators, \re{calendar}, or search engines can help prevent them from hallucinating \cite{schick2023toolformer}. On the other hand, despite their impressive problem-solving abilities, the successful integration of these models into complex systems often requires more than just task understanding - it requires the capacity to manipulate various tools and interact effectively with users.
This is exemplified in systems like AutoGPT \footnote{\url{https://github.com/Significant-Gravitas/Auto-GPT}}, BabyAGI \footnote{\url{https://github.com/yoheinakajima/babyagi}}, and ChatGPT-plugins \footnote{\url{https://openai.com/blog/chatgpt-plugins}}, which leverage LLMs' capabilities beyond merely generating well-written texts and programs. In these systems, LLMs operate as the central controller, manipulating different tools and interacting with humans, thus taking on the role of Artificial Intelligence Agents (AI Agents). In addition to being central planners, LLMs are often used as intermediaries between macro plans and low-level tool calls or as specific tools. As such, LLMs are seen as a crucial approximation of the linguistic world model in real-world systems.

In this paper, we propose a structured framework \re{for LLM-based AI Agents to evaluate the existing LLMs' planning and tool-using ability} and discuss the necessary abilities of such LLM-based AI Agents. Furthermore, we instantiate the framework with different LLMs and evaluate their \textbf{T}ask \textbf{P}lanning and \textbf{T}ool \textbf{U}sage (TPTU) abilities on several tasks. \re{As shown in Figure \ref{fig:intro}, we use the Doraemon as an analogy of our LLM-based agents: Doraemon's magic 4D pocket consists of millions of gadgets (the Tool Set), and Doraemon needs to pick the right tools and solve tasks in a right order.} 
Our main contributions are summarized as follows:

\begin{enumerate}
\item We propose a structured framework tailored for LLM-based AI Agents to evaluate the TPTU abilities of the existing open-source LLMs.

\item We design two distinct types of agents, namely, one-step agent and sequential agent, to execute the inference process of \re{conducting sub-tasks in a once-for-all or sequential manner, respectively}. We provide detailed empirical results and analysis.

\item Our study reveals significant potential in utilizing LLMs for complex tasks. \re{Furthermore, we conclude four following potential weaknesses of LLM-based agents: failing to output in a specific format, struggling to grasp task requirements, over-utilizing one tool, and lack of summary skills. These observations could spark some insights and shed light on the areas that deserve further investigation and improvement.}

\end{enumerate}


\section{Method}
To the best of our knowledge, the study of ``Agent'', ``Autonomous Agent'', ``AI Agent" and ``Multi-Agent'' has been a central part of AI research for decades \cite{jennings1998roadmap, jennings1995applying, franklin1996agent, castelfranchi1998modelling, ferber1999multi, panait2005cooperative}, aimed at understanding and building intelligent and autonomous systems, but there is currently no standardized definition for AI Agents, particularly those that are based on LLMs.

\textbf{In this paper, the Artificial Intelligence Agent (AI Agent) is defined as a program that employs AI techniques to perform tasks that typically require human-like intelligence}. AI Agents can take many forms, from simple chatbots to complex autonomous systems that interact with their environment and make decisions in real-time. They can be trained using a variety of machine learning techniques, including supervised, unsupervised, and reinforcement learning, and can be programmed to perform specific tasks or learn from their experiences in order to improve their performance over time.

\subsection{Agent Framework}

\begin{figure*}[t]
\centering
\includegraphics[width=0.8\textwidth]{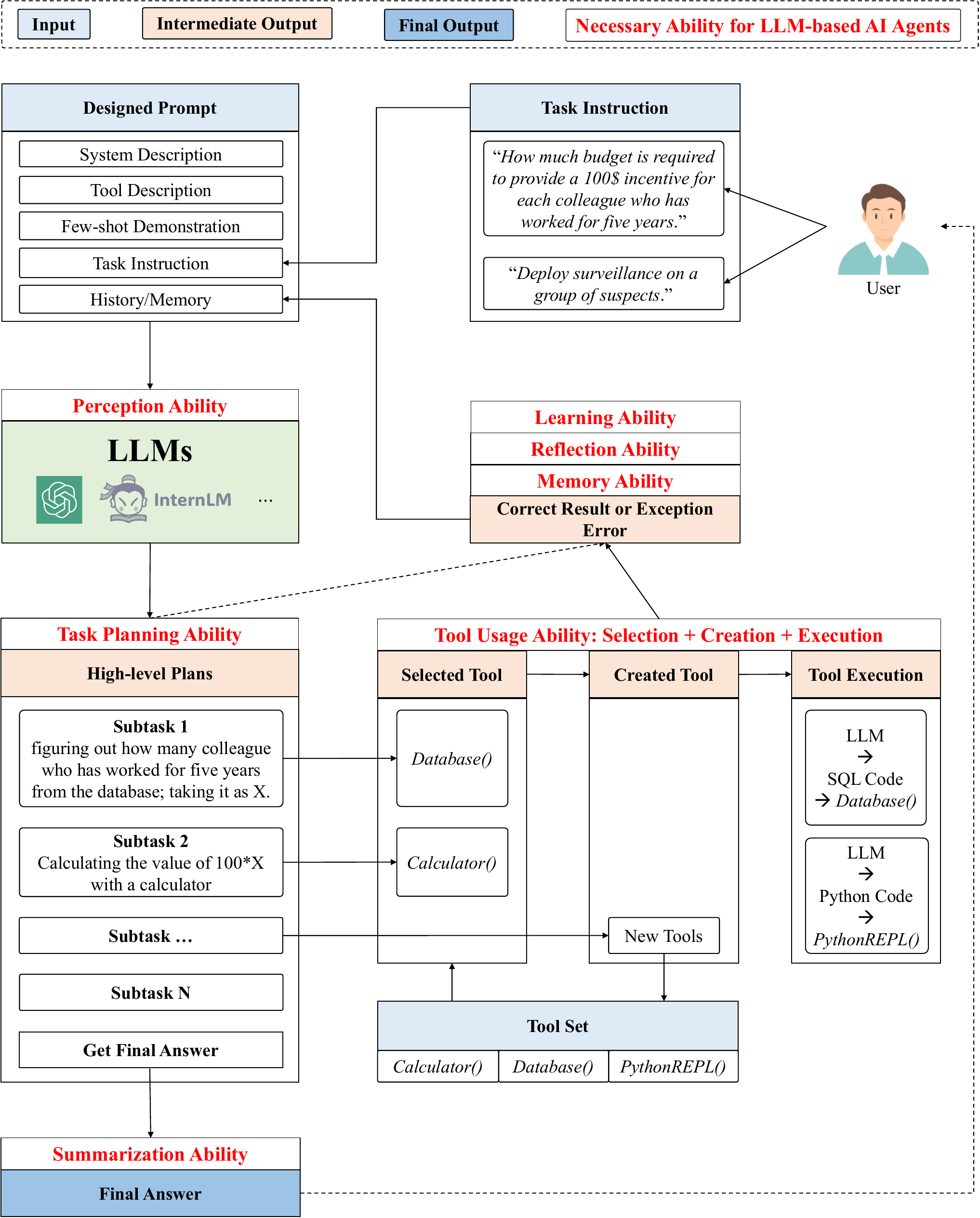}
\caption{The proposed framework for LLM-based AI Agents.}
\label{fig:LLM_AI_Agent}
\end{figure*}

We are particularly interested in the AI Agent that employs the LLM techniques (i.e., LLM-based AI Agent), due to its high efficiency and flexibility in various tasks and domains. Specifically, we design our AI Agent framework with six components as shown in Figure \ref{fig:LLM_AI_Agent}:

\begin{enumerate}
    \item \textbf{Task Instruction}. This is the explicit input of the agent. In practical systems, the task instruction comes from human users of the systems. For example, in a human resources (HR) system, the user may give a task instruction: How much budget is required to provide a 100\$ incentive for each colleague who has worked for five years? In contrast, in a criminal investigation system, the user may give a task instruction: Deploy surveillance on a group of suspects.
    \item \textbf{Designed Prompt}. This is an additional form of input for the agent, derived from tasks that the human users anticipate the AI Agent will complete. Humans can craft specific instructions or demonstrations to steer the LLM-based AI Agents toward generating suitable responses. These guiding inputs could encompass system instructions, tool descriptions, few-shot demonstrations, chat history, or even error output.
    \item \textbf{Tool Set}. It is another input for the agent, which refers to the set of external resources, services, or subsystems that the AI Agent can utilize to aid in its tasks. This could include databases for information retrieval \re{\cite{pourreza2023din}}, APIs for interacting with external systems \re{\cite{schick2023toolformer}}, other AI models specialized for tasks such as image recognition or sentiment analysis \re{\cite{wu2023visual}}, or even non-AI tools and resources such as web scraping tools or data visualization libraries \re{\cite{gorniak2023vizability}}. The toolset expands the capabilities of the AI Agent, enabling it to access and process information beyond its internal knowledge, interact with other systems, or perform specialized tasks that it may not be capable of on its own. For example, an AI Agent might use a weather API to fetch current weather information, or a Python interpreter to solve the mathematical question.
    \item \textbf{LLM}. This is the core component of the system that interprets the task instructions and prompts, interacts with the toolset, and generates the intermediate outputs and final answers. In this context, we utilize publicly available large language models such as ChatGPT, GPT-4 \re{\cite{openai2023gpt4}}, \re{InterLM \cite{2023internlm}}, and others. 
    \item \textbf{Intermediate Output}. This represents the output generated by the LLM-based AI Agent after it processes the task instructions and prompts, and interacts with the toolset. There are three typical intermediate outputs: (1) the high-level plans to fulfill the original user instruction, (2) selected and created tools to fulfill each subtask in the plans, and (3) the results or errors produced after tool execution. The output can be reviewed and refined, either by the AI Agent itself or with human oversight, to ensure it is accurate and meets the requirements of the task instruction.
    \item \textbf{Final Answer}. This is the output that the AI Agent summarizes and provides to the user after all processing (including task planning, tool usage, and possibly error feedback) has been completed.
\end{enumerate}

\subsection{Agent Ability}
To apply LLM-based AI Agents to augment or replace human decision-making in real-world applications, the agents typically require the following abilities:
\begin{enumerate}
    \item \textbf{Perception Ability}: AI Agents must be able to perceive the task instruction from human and system specifications.
    \item \textbf{Task Planing Ability}: AI Agents should have the capacity to create a step-by-step plan for complex task composition based on the perceived instruction and specifications. This usually involves the generation of critical subtask sequences, and the ability to adjust the plan dynamically in response to changes in the task or environment.
    \item \textbf{Tool Usage Ability}: On the one hand, AI Agents should possess the capacity to \textbf{select} a variety of existing tools or resources to execute complex tasks. On the other hand, AI Agents should \textbf{create} new tools by converting the task requirements. This ability enables the AI Agent to extend its capabilities beyond LLM itself and the existing tools by leveraging the vast resources available in the digital world. Finally, AI Agents should be able to \textbf{execute} the selected or created tools for truly grounding the human request based on the resources and constraints of systems. 
    \item \textbf{Learning/Reflection/Memory (from Feedback)}: AI Agents should be capable of learning from feedback, including correct results and exception errors. They should incorporate memory, such as logging or chat history, and reflection to adapt their plans or decisions. This allows the agents to improve their performance and efficiency in task execution continuously.
    \item \textbf{Summarization}: After several rounds of interaction with humans, tools, and systems, AI agents can ultimately complete the original task provided by the users. At this point, AI agents should be able to summarize the interaction history and provide a final answer that is concise and easy to understand for the users.
\end{enumerate}

To endow AI Agents with the aforementioned abilities, some techniques that can be used include chain-of-thought (CoT) and vector databases, as shown in Table \ref{tab:ability_techinique}.

\begin{table}[!ht]
\small
\centering
\caption{A simple illustration of the techniques for endowing the key ability.}
\label{tab:ability_techinique}
\begin{tabular}{ll}
\toprule
\textbf{Ability} & \textbf{Possible Techniques} \\
\midrule
Perception & Multi-input Fusion \\
Task Planing & Zero-shot CoT and  Few-shot CoT \\
\makecell[l]{Tool Usage \\ (Selection/Creation/Execution)} & \makecell[l]{Text Matching/Code Generation/\\Action Grounding} \\
Learning/Reflection/Memory & \makecell[l]{RLHF/Multi-agent Debate/\\Vector Database} \\
Summarization & \makecell[l]{Attention Mechanism and \\Natural Language Generation} \\
\bottomrule
\end{tabular}
\end{table}


\subsection{Agent Design}
\begin{figure*}[ht!]
    \setcounter{subfigure}{0}
	\centering
	\subfigure[One-step Agent (\nameOnestep)]{
		\begin{minipage}[t]{0.83\textwidth}
			\centering
			\includegraphics[width=\textwidth]{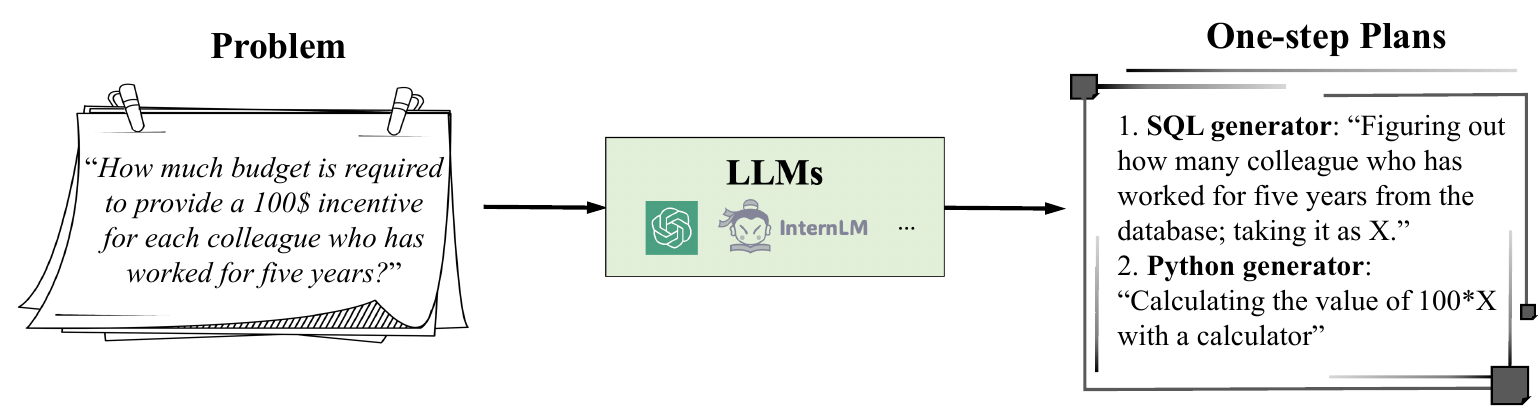}
		\end{minipage}
	}
 
	\subfigure[Sequential Agent (\nameSeq)]{
		\begin{minipage}[t]{0.83\textwidth}
			\centering
			\includegraphics[width=\textwidth]{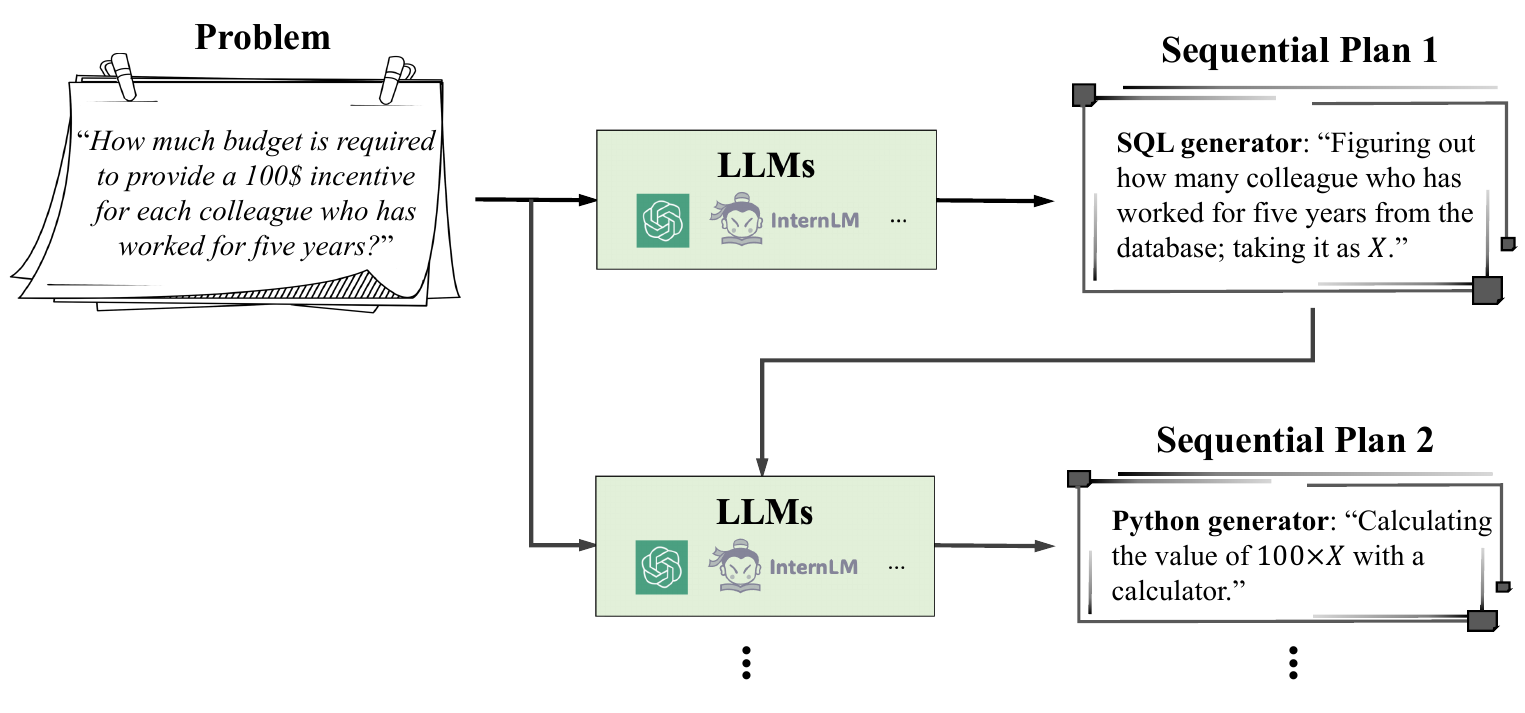}
		\end{minipage}
	}
 \caption{The workflows of the One-step Agent and the Sequential Agent are specifically designed to assess the Task Planning and Tool Usage abilities of LLMs. }
	\label{fig:workflow_agents}\vspace{-10pt}
\end{figure*}

Task planning and tool usage represent the cornerstone of LLM's abilities. Others like perception, learning/reflection/memory (from feedback), and summarization are indeed critical, but they primarily serve to enhance and support these two core competencies.
Therefore, concentrating on these two key competencies - \textbf{T}ask \textbf{P}lanning and \textbf{T}ool \textbf{U}sage (TPTU for short) - we have devised two distinct types of AI agents,
as depicted in Figure~\ref{fig:workflow_agents}:

\begin{itemize}
    \item The first one, named as the \textbf{O}ne-step \textbf{A}gent (\nameOnestep), adopts a global perspective to interpret the original problem, effectively breaking it down into a sequence of sub-tasks in a single instance. This strategy fully harnesses the model's comprehensive understanding capabilities to map out the problem-solving steps for all sub-tasks at once. This method underscores the significance of a holistic understanding and planning of the overall task, albeit it might lack flexibility when dealing with individual sub-tasks.
    
    \item The second type, referred to as the \textbf{S}equential \textbf{A}gent (\nameSeq), emphasizes tackling the current sub-task at hand. Upon successfully resolving the ongoing sub-task, this agent requests the LLMs to provide the succeeding sub-task. This approach enables the model to maintain a clear and concentrated focus throughout the problem-solving journey, tackling issues incrementally. Such a methodology allows for continuous feedback and progress within the confines of addressing a broader problem.
\end{itemize}

These two distinct agent models represent two disparate problem-solving strategies - the \re{one-step and sequential} resolution \footnote{One can also combine the two strategies to design a hierarchical agent, but this is beyond the scope of this paper.}. In our subsequent experiments, we aim to understand their respective strengths and weaknesses and how they can be best utilized to leverage the capabilities of LLMs in real-world problem-solving scenarios.


\section{Evaluation}


We instantiate the proposed LLM-based AI Agent framework (\nameOnestep\;and \nameSeq) with different LLMs and evaluate their performance on typical tasks.





\subsection{Preparations}
Before beginning our evaluation, we first outline the preparations.
We will give detailed descriptions of the datasets, available tools, and popular large language models.

\subsubsection{Datasets}


We first clarify the motivations behind our choice of tools for evaluation. The selection was guided by two primary factors: \textbf{the number of tools} to be evaluated and \textbf{the specific tools} to be included.

Firstly, regarding the number of tools, it is important to state that our proposed evaluation framework is extensible. It can incorporate any number of tools as pluggable components to be managed by the LLM-based AI agents. Secondly, looking at the current work on tool-augmented LLMs, such as T-Bench~\cite{xu2023tool} and ToolBench~\cite{qin2023toolllm}, we see that only a handful of tools are launched and executed in a single scenario. Meanwhile, API-Bank~\cite{li2023api}, in a single scenario, typically dispatches only one API tool and awaits its response. APIBench~\cite{patil2023gorilla} and ToolAlpaca~\cite{tang2023toolalpaca} do not even execute a tool response. Hence, for the sake of simplicity and focus in our evaluation, we have decided to primarily assess two tools (which can be called multiple times) within a single scenario.

Secondly, we also need to decide which specific tools should be used for evaluation. 
Consider a real-world scenario where we pose the question: ``How much budget is required to offer a $\$$100 incentive to each employee who has been with the company for over five years?". To answer this, we first need to retrieve the relevant data from a database, typically using SQL, to find the number of eligible employees. Then, we need to perform a mathematical calculation to estimate the total budget. Such scenarios are quite common in daily life where the formulation and resolution of a question often involve SQL and mathematical tools.

Recognizing the importance of these tools, we have chosen to focus our evaluation on SQL and Python generators, which represent the capabilities of database querying and mathematical computation, respectively. To this end, we have prepared 120 question-answer pairs that vary in complexity. These pairs provide a rigorous assessment of the LLM-based AI agents in understanding, generating, and utilizing these essential tools. For further information on these queries and their corresponding demonstrations, please refer to Appendix~\ref{app:eva_datasets}.

\subsubsection{Tools}

We have defined a total of 12 available tools for the selection of the LLM-based AI agents for evaluation. They are defined as follows:

\begin{itemize}
    \item SQL generator: Given an input question and a database, create a syntactically correct SQLite query statement.
    \item Python generator: Given an input question and some information, generate a syntactically correct Python code.
    \item Weather query tool: Given a location, output the current real-time weather at that location.
    \item Image generator: Given a text description, generate a related image.
    \item Text extractor: Given a link to an image, extract the corresponding text and its position coordinates.
    \item Translator: Given a piece of text, translate it into other languages.
    \item Bing Searcher: Given a piece of text, conduct a search on the Bing browser and return content.
    \item Shell generator: Given an input question and some information, generate a syntactically correct Shell code.
    \item Java generator: Given an input question and some information, generate a syntactically correct Java code.
    \item Wikipedia searcher: Given a piece of text, conduct a search on Wikipedia and return content.
    \item Office software: Given a text description, automatically generate corresponding long documents or spreadsheets or PPTs.
    \item Movie player: Given a movie name, automatically play the corresponding movie resources.
\end{itemize}

\subsubsection{LLMs}
The LLMs evaluated in this paper are listed in Table~\ref{tab:llms-evaluated}, elaborated as follows:
\begin{itemize}
    \item \textbf{GPT} series developed by OpenAI boasts a powerful language model with a vast number of parameters, enabling it to tackle intricate problems efficiently. This paper aims to evaluate the performance of ChatGPT, which balances the performance with costs (the number of OpenAI API calls). 
    \item \textbf{Claude} is committed to maintaining honesty and ensuring user safety, which is developed by Anthropic. With its impressive size, Claude ranks among the largest language models globally and poses a formidable challenge to ChatGPT as a strong competitor.
    \item \textbf{InternLM}, a sophisticated language model developed by Shanghai AI Lab, boasts a multi-round dialogue capability and an impressive ability to comprehend super-long text. This language model is meticulously designed to cater to the nuances of the Chinese language, enabling it to comprehensively understand and effectively process Chinese text. Here, we adopted the version with 120 billion parameters.
    \item \textbf{Ziya} is an expansive and robust pre-training model developed by IDEA, derived from the LLaMa with 13 billion parameters. This comprehensive model exhibits a wide range of capabilities, including translation, programming, and mathematical calculations. Notably, it stands out as a bilingual LLM, highlighting its ability to effectively process and comprehend text in Chinese.
    \item \textbf{ChatGLM}, developed by Tsinghua University, is an open-source dialogue language model that supports bilingual Q\&A in Chinese and English, with a particular focus on Chinese optimization. Built on the General Language Model (GLM) architecture and utilizing model quantization technology, the ChatGLM can be easily deployed on consumer-grade graphics cards, enabling local implementation by users.
    \item \textbf{Chinese-Alpaca-Plus} is achieved by extending LLaMA's existing vocabulary with an additional 20,000 Chinese tokens from Meta AI (formerly known as Facebook AI Research Laboratory). In this version, we use a model with 33 billion parameters. The training text has been expanded to 120GB, and the fine-tuning instruction data has been increased to 4.3M.
\end{itemize}

\begin{table*}[ht!]
    \centering
    \caption{The LLMs evaluated in this paper.}
    \label{tab:llms-evaluated}
    \begin{tabular}{cccc}
    \toprule
        \textbf{Organization} & \textbf{Model Name} & \textbf{Model Parameters} \\ 
        \midrule
		OpenAI & \href{https://chat.openai.com/}{ChatGPT}\cite{ouyang2022training} & 200B \\ 
        \hline
            Anthropic & \href{https://claude.ai}{Claude}\cite{bai2022constitutional} & >52B \\ 
        \hline
            Shanghai AI Lab & InternLM & 120B \\ 
        \hline
            IDEA & \href{https://github.com/IDEA-CCNL/Fengshenbang-LM}{Ziya-13B} & 13B \\ 
        \hline
            Tsinghua University & \href{https://github.com/THUDM/GLM-130B}{ChatGLM-130B}\cite{zeng2022glm} & 130B \\ 
        \hline
            - & \href{https://github.com/ymcui/Chinese-LLaMA-Alpaca/}{Chinese-Alpaca-Plus-33B}\cite{touvron2023llama,cui2023efficient_alpaca} & 33B \\ 
    \bottomrule
    \end{tabular}
\end{table*}

\subsection{Evaluation on Task Planning Ability}

In this section, to evaluate the planning capabilities of the LLM-based AI agents, we have structured the evaluations as follows.

For TPTU-OA, we begin by examining the agents' ability to plan the order of tool use. This is followed by an evaluation of the agents' capacity to not only plan the sequence of tools but also the corresponding subtask descriptions. Subsequently, we conduct a specialized planning evaluation where the agents must generate multiple sequences of key-value pairs of the form \{tool: subtask description\} in complex problem teardowns.
Moreover, we expand the toolset with additional, unrelated tools to further challenge and reassess the planning ability of the LLM-based AI agents.

For TPTU-SA, we follow the regime that the agent should generate multiple sequences of key-value pairs of the form \{tool: subtask description\} for evaluation.

\subsubsection{\nameOnestep: Tool Order Planning}
Here, we utilize two kinds of tools for problem-solving: the SQL generator, which retrieves data from databases, and the Python generator, adept at addressing mathematical questions. 

To validate the capacity of the LLM-based AI agents to strategically plan for the tool order, we designed the prompt as shown in Figure~\ref{tab:prompt_plan_tool_name} of Appendix~\ref{app:prompts}.
This design is motivated by the goal to assess the ability of LLM-based AI agents to understand complex problems, subsequently decomposing them into a sequence of simpler tasks executed by appropriately selected tools.
Specifically, we require the LLM-based AI agent to follow our instructions, select tools from our pre-defined tool set with detailed function descriptions, conform to the given format strictly, and understand the demonstrations to learn from them.

Upon feeding these prompts into the LLM-based AI agents under evaluation, we obtained the following accuracy rates for the tool planning, as shown in Table~\ref{tab:res_plan_tool_name}.


\begin{table*}[ht!]
    \centering
    \caption{The evaluation results for the planning of tool order generation.}
    \label{tab:res_plan_tool_name}
    \begin{tabular}{cccccc}
    \toprule
        \textbf{Model} & ChatGPT  & Claude   & Ziya   \\ 
        \cline{2-4}
        \textbf{Accuracy} & 100\%  & 100\%  & 45\%     \\
        \hline
        \textbf{Model}  & ChatGLM & Chinese-Alpaca-Plus &  InternLM  \\ 
        \cline{2-4}
        \textbf{Accuracy} & 45\%  & 20\%  & 80\%        \\
    \bottomrule
    \end{tabular}
\end{table*}


The results of our experiments indicate that models, notably Ziya and ChatGLM, frequently grapple with the generation of lists in the correct format. For other models, the predominant challenges lie in generating tools in the correct sequence or in the occasional omission of necessary tools. Nonetheless, the issue of parsing list formats is generally negligible.

These findings suggest that the majority of LLM-based AI agents possess a fundamental capability to analyze the tool needs of a given problem and understand its task requirements. To further explore whether these LLM-based AI agents can effectively break down the original problem into sub-tasks, we proceed to the following section.

\subsubsection{\nameOnestep: Tool Order Planning and Subtask Description Generation}

Simply planning the order of tool usage is not sufficient to fully address a problem. To truly solve it, we need to provide a guide or instructions for the usage of each tool, that is, a decomposed subtask description. 
Therefore, we can decompose the original complex problem into two separate sequences. One sequence represents the order in which the tools are utilized, while the other sequence corresponds to the subtask descriptions that each tool in the tool sequence aims to resolve. A problem is only truly solved when both the tool and subtask description sequences have been successfully planned. 
In order to verify whether LLM-based AI agents truly have the ability to solve complex problems, we designed a new prompt as shown in Figure~\ref{tab:prompt_plan_tool_name_query} of Appendix~\ref{app:prompts}.
The main improvement is to plan the corresponding subtask description for each tool after the tool planning is completed.


\begin{table*}[ht!]
    \centering
    \caption{The evaluation results for the planning of tool order and subtask description generation.}
    \label{tab:res_plan_tool_name_query}
    \begin{tabular}{cccccc}
    \toprule
        \textbf{Model} & ChatGPT  & Claude   & Ziya   \\ 
        \cline{2-4}
        \textbf{Accuracy} & 55\%  & 15\%   & 10\%     \\
        \hline
        \textbf{Model}  & ChatGLM & Chinese-Alpaca-Plus &  InternLM  \\ 
        \cline{2-4}
        \textbf{Accuracy} & 10\%  & 0\%  & 45\%         \\
    \bottomrule
    \end{tabular}
\end{table*}

After feeding the prompt to these LLM-based AI agents, we get results shown in Table~\ref{tab:res_plan_tool_name_query}.


Although the generation of tool sequences and their corresponding subtask descriptions might be an effective way to problem-solving, there is a significant decrease in accuracy for all LLMs as can be seen. We hypothesize that there are a few potential drawbacks to this method:
\begin{enumerate}
    \item \textbf{Difficulty in Error Tracking and Debugging}. Generating the complete tool and subtask sequences may make it more challenging to track and debug errors. If an error arises within the sequence, it might require a total regeneration instead of a simple modification or repair to the erroneous part.
    \item \textbf{Tool-Subtask Pairing Issue}. If all tool sequences and subtask descriptions are generated independently, there's an inherent risk of misalignment between the tools and their corresponding subtasks. This could potentially lead to an improper pairing, which, in turn, could result in a flawed or ineffective solution that fails to appropriately resolve the given problem.
    \item \textbf{Lack of Flexibility}. The approach may lack this flexibility when facing complex problems requiring adjustments to the tool or subtask sequence.
    \item \textbf{Dependency on Global Information}. Generating the entire tool and subtask sequences requires a global understanding and planning of the entire problem. However, in some instances, certain parts of the problem might not be clear at the early stages of problem-solving, which could pose challenges within this framework.
\end{enumerate}


\subsubsection{\nameOnestep: The Planning of Tool-Subtask Pair}

To mitigate the aforementioned issue, we propose a novel approach to foster flexible problem-solving with the LLM-based AI agent. We prompt the agent to generate multiple sequences, each consisting of a key-value pair in the format of \{tool: subtask description\} that associates a tool with its respective subtask description. This allows us to simultaneously plan the tool choice and subtask without the risk of improper matching. Moreover, it offers the flexibility to update the planned sequences in real-time based on evolving problem feedback, enhancing adaptability and efficiency when addressing complex tasks.

With this consideration, we have designed a unique prompt that encourages this advanced problem-solving strategy. In the following section, we delve into the specifics of this prompt design in Figure~\ref{tab:prompt_plan_oneshot} of Appendix~\ref{app:prompts}. 
The key improvement in this prompt is its directive for the LLM-based AI agents to stringently adhere to the predefined dictionary format. To facilitate this, we offer several demonstrations in our desired format, serving as references for the language model to follow.



\begin{table*}[ht!]
    \centering
    \caption{The evaluation results for the planning of Tool-Subtask pair.}
    \label{tab:res_plan_one_shot}
    \begin{tabular}{cccccc}
    \toprule
    \textbf{Model} & ChatGPT  & Claude   & Ziya   \\ 
        \cline{2-4}
        \textbf{Accuracy} & 75\%  & 90\%   & 20\%     \\
        \hline
        \textbf{Model}  & ChatGLM & Chinese-Alpaca-Plus &  InternLM  \\ 
        \cline{2-4}
        \textbf{Accuracy} & 0\%  & 5\%  & 55\%        \\
    \bottomrule
    \end{tabular}
\end{table*}

After feeding the prompt to these LLM-based AI agents, we get results shown in Table~\ref{tab:res_plan_one_shot}. 



Analyzing the results from Tables~\ref{tab:res_plan_tool_name_query} and~\ref{tab:res_plan_one_shot}, we observe a marked improvement of 52.9\% when the tool-subtask pairs are generated in a unified format compared to separate generation of tools and subtasks.

This significant performance enhancement can likely be attributed to the close coupling between tools and their associated subtasks in our unified generation strategy. When tools and subtasks are generated separately, there is a potential disconnect or lack of coherence between the two, which could lead to less accurate or efficient solutions. In contrast, by generating tool-subtask pairs together, we ensure that each tool is directly tied to its relevant subtask, leading to a more coordinated and effective problem-solving approach. This might explain the observed increase in overall performance.


\subsubsection{\nameOnestep: The Planning of Tool-Subtask Pair with Unrelated Tools}

So far, our analysis and evaluation have been primarily focused on the LLM-based AI agents' proficiency in planning with specific tools. However, we are also interested in how it would perform when faced with many irrelevant or similar tools. Therefore, for a more comprehensive assessment, we expanded the prompt in Table~\ref{tab:prompt_plan_oneshot} to include an additional ten unrelated tools, as illustrated in Figure~\ref{tab:prompt_plan_others} of Appendix~\ref{app:prompts}.



\begin{table*}[ht!]
    \centering
    \caption{The evaluation results for the planning of Tool-Subtask pair with unrelated tools.}
    \label{tab:res_plan_one_shot_with_other_tools}
    \begin{tabular}{cccccc}
    \toprule
        \textbf{Model} & ChatGPT  & Claude   & Ziya   \\ 
        \cline{2-4}
        \textbf{Accuracy} & 70\%  & 90\%    & 10\%     \\
        \hline
        \textbf{Model}  & ChatGLM & Chinese-Alpaca-Plus &  InternLM  \\ 
        \cline{2-4}
        \textbf{Accuracy} & 0\%  & 5\%  & 50\%       \\
    \bottomrule
    \end{tabular}
\end{table*}

After feeding the prompt to these LLM-based AI agents, we get results shown in Table~\ref{tab:res_plan_one_shot_with_other_tools}. 
The results from our expanded evaluation demonstrate that even when presented with irrelevant or similar tools and descriptions, LLM-based AI agents consistently avoid selecting these unrelated tools (i.e., the accuracy has remained unchanged or exhibited only a marginal decrease compared with Table~\ref{tab:res_plan_one_shot}). This outcome indicates the effectiveness of our designed prompt, which successfully guides the LLM-based agents to understand the appropriate tool sequence for complex problem decomposition.

This observation reinforces the notion that a well-structured and informative prompt can efficiently guide AI agents to understand the core essence of the problem, thereby enabling them to sift through irrelevant information and focus on key tasks. This successful discrimination against unrelated tools also points towards the models' ability to understand the specific context of a problem and select the appropriate tools, thereby enhancing the overall problem-solving process.
\subsubsection{\nameSeq: The Planning of Tool-Subtask Pair Generation}

Upon identifying the drawbacks of first generating a list of tools and then generating corresponding subtask descriptions, we decided to focus subsequent tests on the generation of tool-subtask pairs. Consequently, in this section, we evaluate the capability of TPTU-SA to generate these tool-subtask pairs.

To achieve the goal of recursively generating tool-subtask pairs, we have designed prompts as illustrated in Figure~\ref{tab:prompt_sa_planning} of Appendix~\ref{app:prompts}.

\begin{table*}[ht!]
    \centering
    \caption{The evaluation results for the planning of Tool-Subtask with the sequential agent.}
    \label{tab:res_plan_sequential_agent}
    \begin{tabular}{cccc}
    \toprule
        \textbf{Model} & ChatGPT  & Claude   & Ziya   \\ 
        \cline{2-4}
        \textbf{Accuracy} & 80\%  & 100\%    & 10\%     \\
        \hline
        \textbf{Model}  & ChatGLM & Chinese-Alpaca-Plus &  InternLM  \\ 
        \cline{2-4}
        \textbf{Accuracy} & 0\%  & 0\%  & 65\%        \\
    \bottomrule
    \end{tabular}
\end{table*}

The evaluation results are shown in Table~\ref{tab:res_plan_sequential_agent}. Compared with results shown in Table~\ref{tab:res_plan_one_shot}, TPTU-SA generally performs better than TPTU-OA especially for high--performing LLMs (e.g., ChatGPT, Claude and InternLM). We propose the following potential reasons for this observation:

\begin{enumerate}
    \item \textbf{Sequentiality Mimics Human Problem-Solving}: In real-world scenarios, humans tend to solve complex problems by breaking them down into smaller, manageable subtasks which are often handled sequentially. Sequential agents are designed to mimic this step-by-step approach, which might inherently suit complex problem-solving better.
    \item \textbf{Richer Contextual Understanding}: Sequential agents are exposed to the outcome of each previous subtask before moving on to the next one. This iterative process could facilitate a richer understanding of the problem context, enabling more accurate task planning and tool usage.
    \item \textbf{Flexibility in Task Management}: In comparison to one-step agents, sequential agents might have more flexibility in managing tasks. They have the opportunity to correct errors or adjust their strategy after each step, which can lead to improved overall performance.
    \item \textbf{Improved Learning From History}: The sequential process provides a history of actions and results which can be beneficial in learning. The agent can use this history to make better predictions about what tool to use next or what subtask to tackle, leading to more accurate and efficient problem-solving.
\end{enumerate}

These points of analysis suggest that the structure and operation of sequential agents inherently confer certain advantages in complex problem-solving scenarios, leading to their superior performance.

\subsection{Evaluation on Tool Usage Ability}


Before evaluating the end-to-end multi-tool usage ability of LLM-based AI agents, we first evaluate the effectiveness of single-tool usage for SQL generation and mathematical code generation.

Subsequently, to assess the end-to-end performance of LLMs across various tools, two types of agents (TPTU-OA and TPTU-SA) were developed and several LLMs were subjected to testing under these agents. The role of the agents is to break down complex questions into simpler sub-questions and plan corresponding tools to solve them, based on the available toolset and corresponding tool descriptions.



\subsubsection{The effectiveness of Single Tool Usage}
\label{sec:eva_tool_creation}

Our aim is to systematically assess how effectively these models can use various tools, focusing on their proficiency with SQL and other coding languages.

\paragraph{\textbf{The Effectiveness of simple SQL Creation}}

Using the schemas provided in Table~\ref{tab:schema_simple1} and Table~\ref{tab:schema_simple2}, we construct questions similar to those, and refer readers to Appendix~\ref{app:eva_datasets}. These questions are posed to various LLMs using our specifically designed prompts in Appendix~\ref{app:prompts}.


Following the tailored prompts, the LLMs are evaluated based on their responses to the presented queries. The results of this comprehensive assessment are compiled and exhibited in Figure~\ref{tab:evaluated_saimple_sql}.


\begin{table*}[ht!]
    \centering
    \caption{The evaluation results for simple SQL questions.}
    \label{tab:evaluated_saimple_sql}
    \begin{tabular}{cccccc}
    \toprule
    \textbf{Model} & ChatGPT  & Claude    & Ziya   \\ 
        \cline{2-4}
        \textbf{Accuracy} & 90\%  & 100\%     & 50\%     \\
        \hline
        \textbf{Model}  & ChatGLM & Chinese-Alpaca-Plus &  InternLM   \\ 
        \cline{2-4}
        \textbf{Accuracy} & 30\%  & 20\%  & 90\%       \\
    \bottomrule
    \end{tabular}
\end{table*}

This verifies the capabilities of each LLM in handling varying simple single-table SQL queries, thus providing a basis for comparison and analysis. 

\paragraph{\textbf{The Effectiveness of Complex Nested SQL Creation}}

Using the schemas provided in Table~\ref{tab:schema_nest1},~\ref{tab:schema_nest2},~\ref{tab:schema_nest3}, and~\ref{tab:schema_nest4}, we construct questions similar to those, and refer readers to Appendix~\ref{app:eva_datasets}. 
For complex nested SQL questions, to further verify the SQL tool creation capability of LLMs, we have designed two types of prompts. One is the direct-guidance type, which explicitly informs the model that it needs to generate nested SQL query statements, as shown in Figure~\ref{tab:prompt_nested_sql} in Appendix~\ref{app:prompts}.

The other is based on the Chain-of-Thought (CoT)~\cite{CoT2022} approach, which leverages the model's ability to reason step by step to comprehend and craft SQL tools, and the prompt is shown in Figure~\ref{tab:prompt_nested_sql_cot} in Appendix~\ref{app:prompts}.
This method guides the model to sequentially generate SQL query clauses based on the problem context, thus breaking down the complex query generation task into smaller and manageable subtasks. This approach provides the model with a structured way to handle complex SQL tasks and showcases its capacity to engage in incremental reasoning and problem-solving.

The design of these two types of prompts serves as the backbone of our evaluation for complex nested SQL questions. While the direct-guidance approach focuses on testing the model's raw ability to generate SQL queries when explicitly instructed, the CoT-based approach evaluates a more nuanced capability: the model's reasoning and problem-solving skills in a step-by-step manner. Both these methods present unique challenges and offer valuable insights into the strengths and potential areas of improvement for the large language model's SQL tool generation ability.
Subsequently, we will explore these two dimensions based on our experimental evaluations shown in Table~\ref{tab:evaluated_complex_nested_sql}.


\begin{table*}[ht!]
    \centering
    \caption{The evaluation results for complex nested SQL questions.}
    \label{tab:evaluated_complex_nested_sql}
    \begin{tabular}{cccc}
    \toprule
        \textbf{Model} & ChatGPT  & Claude   & Ziya    \\ 
        \cline{2-4}
        \textbf{Direct-based} & 80\%   & 100\%     & 50\%   \\
        \textbf{CoT-based} & 80\%  & 100\%    & 40\%   \\
     \midrule
        \textbf{Model} & ChatGLM & Chinese-Alpaca-Plus & InternLM      \\ 
        \cline{2-4}
         \textbf{Direct-based} & 60\%  & 0\%  & 60\%     \\
         \textbf{CoT-based} & 70\%  & 0\%  & 50\%        \\
    \bottomrule
    \end{tabular}
\end{table*}


From the above results in Table~\ref{tab:evaluated_complex_nested_sql}, it is clear that different models possess varying levels of proficiency in handling complex nested SQL tasks. Some models, like Claude, exhibit a robust capability in SQL generation, no matter whether the approach is direct or CoT-based. Most of these models demonstrate the SQL tool usage capability.

Specifically, some models such as ChatGLM show a distinct preference for the CoT-based approach, their performance improves when problems are broken down into smaller, manageable sub-tasks. This suggests that these models may have a stronger ability in sequential problem-solving and benefit more from step-by-step guidance. Conversely, models like Ziya and InternLM show a drop in performance when tasks are guided in the CoT-based format. This might indicate challenges in managing dependencies between sub-tasks or handling the continuity in sequential problem-solving. Lastly, Chinese-Alpaca-Plus shows significant room for improvement in complex SQL generation tasks. This shows that not all models are equally suited to handle advanced problem-solving involving nested SQL queries.

Overall, these findings underscore the importance of tailoring evaluation and training methodologies to the individual strengths and weaknesses of each model. By adopting this approach, we can better understand the performance variations across different models and provide targeted improvements to enhance their problem-solving abilities. Furthermore, this analysis highlights the potential of LLM-based agents in real-world applications, and the need to push their boundaries through continued research and development.

\paragraph{\textbf{The Effectiveness of Mathematical Code Creation}}

Following our evaluation of the LLM's proficiency in creating complex SQL queries, we now shift our focus to another tool creation: the creation of mathematical code. To the best of our knowledge, while large language models possess significant capabilities, they often fall short of providing highly accurate solutions to mathematical problems. 
Guiding these LLMs to generate mathematical code, and subsequently leveraging external tools to execute and derive the solutions, could significantly enhance their ability to tackle mathematical challenges.

In the upcoming section, we will conduct a detailed evaluation of guiding these LLMs to generate mathematical code.
We aim to shed light on the true capability of these models in generating mathematical code and to elucidate the extent to which they can be utilized to aid in mathematical problem-solving.
The prompt about how to guide LLMs is shown in Figure~\ref{tab:prompt_math_qa} in Appendix~\ref{app:prompts}.


\begin{table*}[ht!]
    \centering
    \caption{The evaluation results for mathematical questions.}
    \label{tab:evaluated_math_qa}
    \begin{tabular}{cccccc}
    \toprule
    \textbf{Model} & ChatGPT  & Claude   & Ziya   \\ 
        \cline{2-4}
        \textbf{Accuracy} & 90\%  & 85\%   & 50\%     \\
        \hline
        \textbf{Model}  & ChatGLM & Chinese-Alpaca-Plus &  InternLM   \\ 
        \cline{2-4}
        \textbf{Accuracy} & 0\%  & 55\%  & 95\%        \\
    \bottomrule
    \end{tabular}
\end{table*}



The results shown in Table~\ref{tab:evaluated_math_qa} indicate that the capabilities of LLM-based agents to generate mathematical code vary considerably. High-performing models like ChatGPT, Claude, and InternLM display excellent proficiency, suggesting their potent ability to solve complex mathematical tasks.
Middle-tier models, such as Ziya, show moderate success, indicating the potential for improvement and adaptability with the right training and optimization. Surprisingly, Alpaca demonstrated a notable proficiency in mathematical tasks, despite its poor performance in SQL generation, suggesting a possible inclination towards mathematical problems.
In contrast, ChatGLM struggles significantly with mathematical code generation, underlining a potential weak spot in its capabilities and the need for focused improvement in this area.

Overall, these results underscore the task-dependent nature of LLMs' capabilities and highlight the importance of recognizing their individual strengths and weaknesses for optimal model guidance and enhanced problem-solving.

\subsubsection{\nameOnestep and \nameSeq: Tool Usage for Multiple Tools}

We now aim to utilize the one-step agent and sequential agent, which we designed, to conduct an evaluation involving multiple tools. Corresponding prompts for each agent type have been crafted and are presented in Figure~\ref{tab:prompt_end2end_onestep} and Figure~\ref{tab:prompt_end2end_sequential} of Appendix~\ref{app:prompts}, respectively.

In this phase of the evaluation, we need to automatically invoke the respective tools through code and produce the results. Given that user interface-based LLMs lack the capability to call external tools, we will only utilize the following four API-based LLMs (ChatGPT, Ziya, Chinese-Alpaca, and InternLM) for this comprehensive evaluation of external tool usage ability.




\begin{table*}[ht!]
    \centering
    \caption{The evaluation results for end-to-end ability of multiple tools.}
    \label{tab:evaluated_end2end_multi_tool}
    \begin{tabular}{ccccc}
    \toprule
        \textbf{Model} & ChatGPT & Ziya & Chinese-Alpaca-Plus & InternLM   \\ 
        \cline{2-5}
        \textbf{\nameOnestep}   & 50\%  & 0\%  & 0\% & 15\% \\    
        \textbf{\nameSeq}     & 55\%  & 0\%  &  0\%  & 20\%\\    
    \bottomrule
    \end{tabular}
\end{table*}

With agents mentioned above, the final results are presented in Table \ref{tab:evaluated_end2end_multi_tool}.
The evaluation results demonstrate varying levels of task planning and tool usage capabilities among the four API-based LLMs. In the TPTU-OA evaluation, ChatGPT achieved a performance rate of 50\%, significantly outperforming the other models, with InternLM at 15\%, while both Ziya and Chinese-Alpaca did not manage to complete any tasks successfully, resulting in a score of 0\%.
In the TPTU-SA evaluation, an overall slight improvement was observed. ChatGPT maintained its leading position, with a slightly improved performance rate of 55\%. InternLM also exhibited better performance, achieving a score of 20\%, whereas Ziya and Chinese-Alpaca-Plus again failed to register any successful task completion.

These results reflect a notable discrepancy in the performance of LLMs when it comes to using external tools. ChatGPT and InternLM have demonstrated some ability to navigate these tasks, but their performance rates suggest there is significant room for improvement. Ziya and Chinese-Alpaca-Plus' performance indicates a struggle to effectively utilize external tools in their current state.

The differential performance between the TPTU-OA and TPTU-SA evaluation hints at the possible impact of the agent design on the LLMs' task execution ability. In particular, the performance increase under the sequential agent framework suggests that breaking down tasks into sequential steps might help LLM-based AI agents better utilize external tools. This insight could prove valuable in future improvements and developments of LLM-based AI agents. 
However, even with this approach, it is clear that LLM-based AI agents are far from perfect when it comes to effectively using external tools for complex tasks. This finding underlines the importance of further investigation and improvement in this domain.

\subsection{Insightful Observations}

Upon closer observation of our experimental results, we have identified several phenomena that deserved further exploration. These findings serve to broaden our understanding of LLM-based agents' behavior and capabilities and provide essential insights that could shape future research in this field. In the following, we will dissect these phenomena as shown in Figure \ref{fig:issue1} - \ref{fig:issue4}, casting light on the weaknesses of LLM-based agents in the context of task planning and tool usage.

\begin{enumerate}
    \item \textbf{Misunderstanding Output Formats}: LLMs frequently encounter difficulty when output is required in specific formats such as lists or dictionaries. One such example includes inconsistencies between the number of tools and corresponding subtasks, leading to formatting issues that hinder the correct execution of tasks.
    
\begin{figure}[ht!]
\centering
\includegraphics[width=0.9\columnwidth]{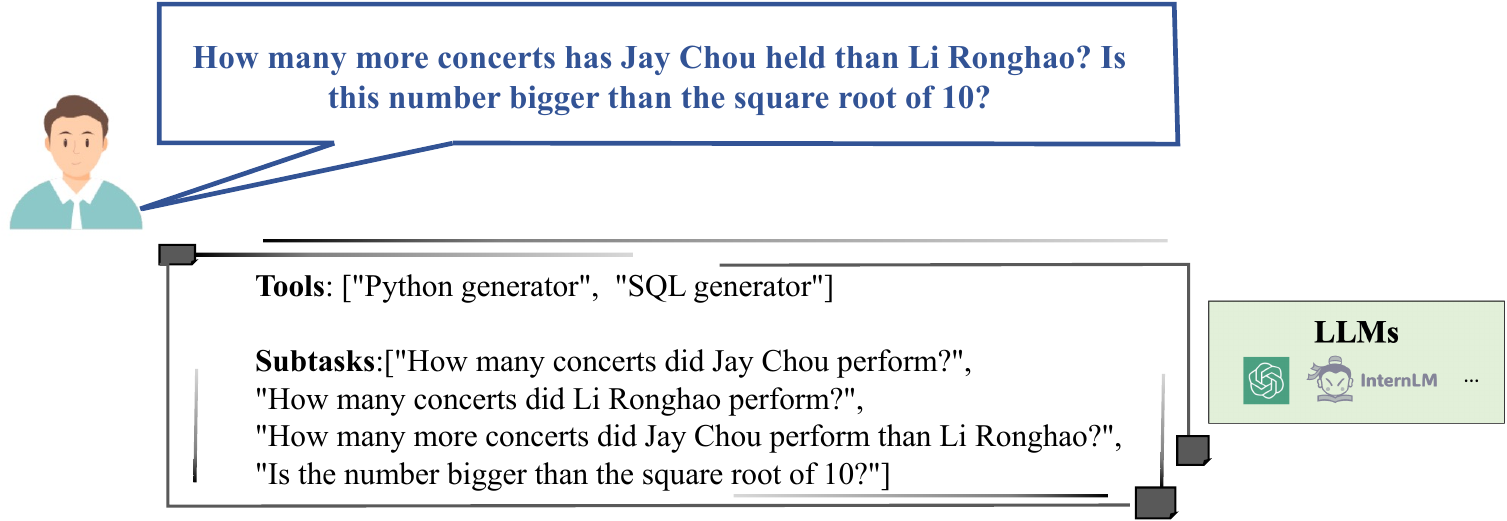}
\caption{\textbf{Issue-1}: Inconsistencies between the number of tools and corresponding subtasks.}
\label{fig:issue1}
\end{figure}

    \item \textbf{Struggling to Grasp Task Requirements}: LLMs might incorrectly disintegrate subproblems or apply unsuitable tools to carry out the subproblem. For example, an LLM might attempt to solve a purely mathematical problem by employing an SQL tool or could misunderstand similar terms like cube extraction and cube roots.
    
\begin{figure}[ht!]
\centering
\includegraphics[width=0.9\columnwidth]{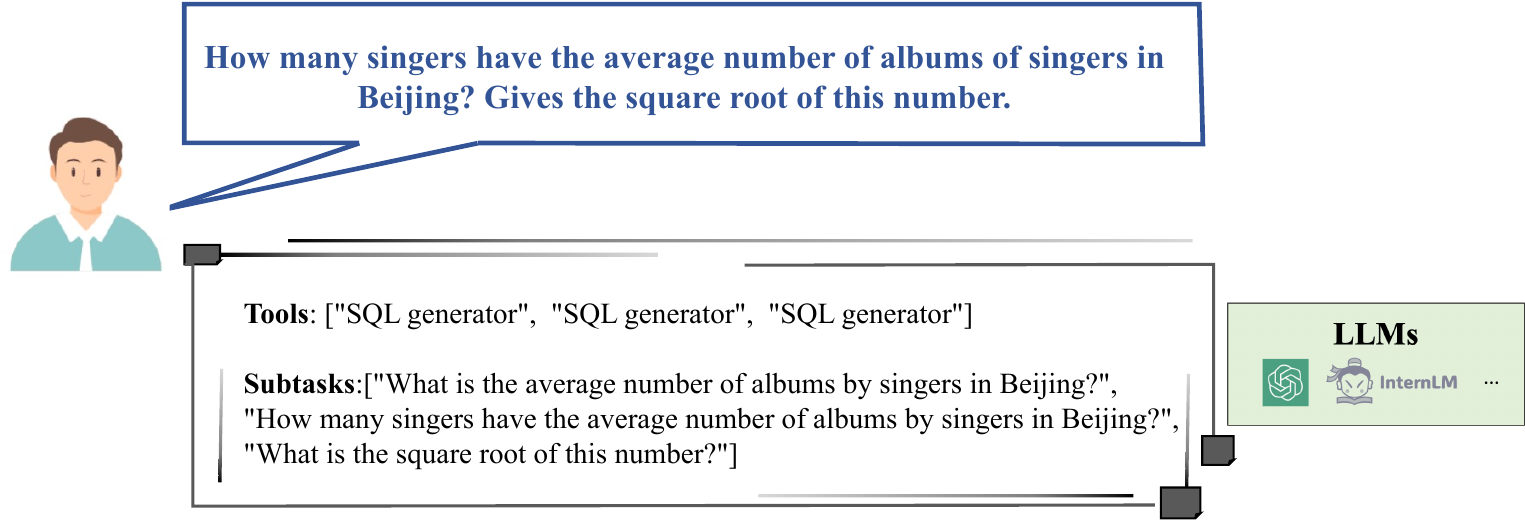}
\caption{\textbf{Issue-2}:Solve a purely mathematical problem by employing a
SQL generator.}
\label{fig:issue2}
\end{figure}
    
    \item \textbf{Endless Extensions}: LLMs tend to overutilize a particular tool, even in instances where a single use would suffice for the correct result. This issue can lead to extended and nonsensical planning, where the same subtask is repeatedly solved.

\begin{figure}[ht!]
\centering
\includegraphics[width=0.9\columnwidth]{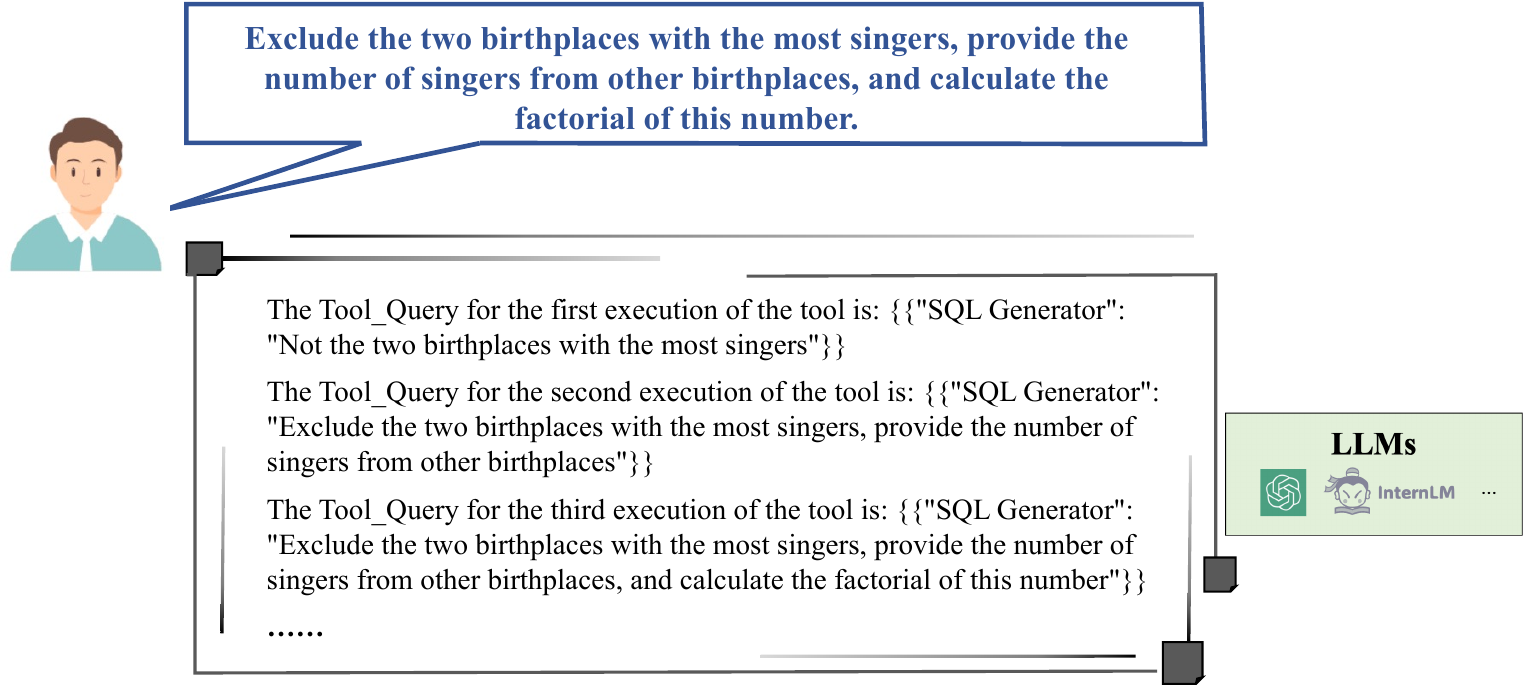}
\caption{\textbf{Issue-3}: Unnecessary repetition of subtasks.}
\label{fig:issue3}
\end{figure}

    \item \textbf{Lack of Summary Skills}: LLMs do not take into account the responses to subproblems, relying instead on their internalized knowledge to generate the final answer. This may lead to a scenario where the final response only addresses a portion of the original query.
\begin{figure}[ht!]
\centering
\includegraphics[width=0.9\columnwidth]{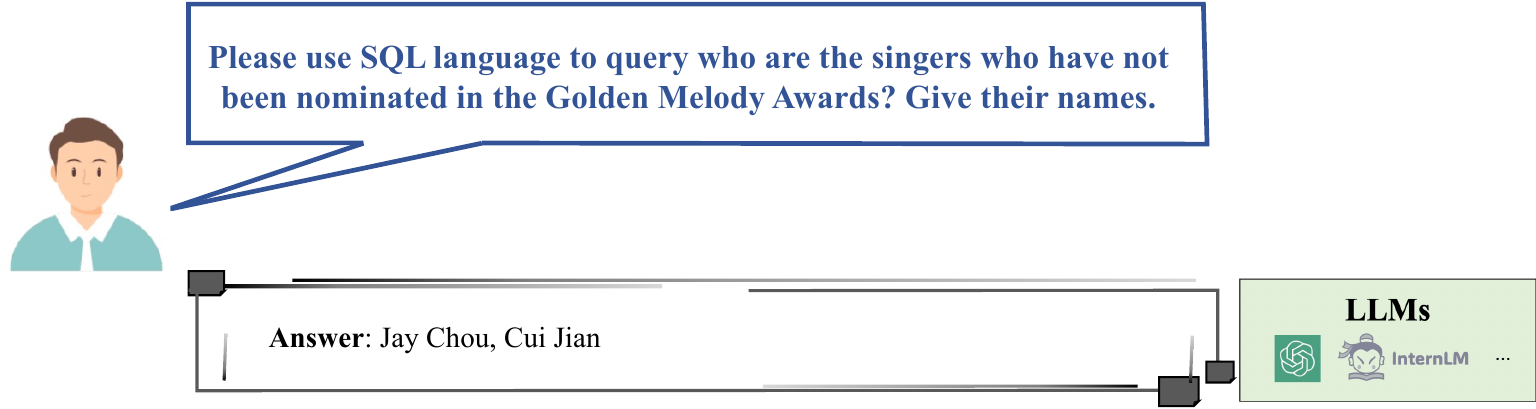}
\caption{\textbf{Issue-4}: Answering questions using common sense instead of generating code.}
\label{fig:issue4}
\end{figure}
\end{enumerate}

By identifying and addressing these common issues, we stand a better chance at improving and refining LLMs, thereby unlocking their full potential.


\section{Related Work}

The remarkable capacity for usage and  creation of tools have facilitated the transcendence of our innate physical and cognitive constraints, 
thereby profoundly advancing the progress and prosperity of human civilization and society. 
The swift advancement of LLM has rendered it feasible to use and create tools like humans. 
The integration of specialized tools with LLM has unlocked substantial potential in addressing intricate tasks. 
In this section, we offer a concise synopsis of the relevant research pertaining to tool learning based on LLMs.

\subsection{Tool Usage}

The initial advancements in tool learning have been constrained by the capabilities of artificial intelligence (AI) models. \cite{bommasani2021opportunities}
Traditional deep learning approaches exhibit limitations in terms of comprehension of tool functionality and user intentions, and common sense reasoning abilities.
Consequently, these limitations directly result in a notable decline in the stability and precision of tool learning methodologies.
Recently, the advent of LLM has marked a pivotal juncture in the realm of tool learning. LLMs encompass a broad spectrum of common sense cognitive capabilities and exhibit remarkable proficiencies in natural language processing, reasoning, and interactive decision-making \cite{mosbach2023few, yang2023harnessing, zhang2023one, yu2023nature, wang2023interactive}. 
These attributes furnish indispensable prerequisites for LLMs to comprehend user intentions and effectively employ tools in tackling intricate tasks \cite{qin2023tool}. 
Simultaneously, the advancement of fine-tuning \cite{yu2022survey, hu2021lora, houlsby2019parameter, li2021prefix, liu2021gpt} and in-context learning \cite{yao2022react,khot2022decomposed} technology has offered robust support to LLM in addressing increasingly intricate challenges.
In addition, tool usage can mitigate the inherent limitations of LLMs, encompassing the acquisition of up-to-date information from real-world events, refined mathematical computational abilities, and the mitigation of potential hallucinatory phenomena. \cite{mialon2023augmented}

Within the realm of embodied intelligence \cite{duan2022survey, savva2019habitat, franklin1997autonomous}, 
LLM engages in direct interactions with tangible tools like robots in order to enhance their cognitive abilities, 
optimize work productivity, and expand functional capacities. 
LLM possesses the capability to automatically devise action steps based on user intentions, enabling the guidance of robots in the completion of tasks \cite{zhang2023lp, shah2023lm, brohan2023can, huang2022inner, chen2023open, driess2023palm, wake2023chatgpt, rana2023sayplan, song2022llm}, or alternatively, to directly generate underlying code that can be executed by robots \cite{brohan2022rt, stone2023open, reed2022generalist, vemprala2023chatgpt, liang2023code}.
Palm-E \cite{driess2023palm} introduced a multimodal language model which seamlessly integrates sensor data into its framework, 
enabling efficient planning of robot actions and task completion. 
Code as Policies (CaP) \cite{liang2023code} facilitates the transformation of natural language instructions into code fragments that can be directly compiled and executed on robots. 
As for Inner Monologue \cite{huang2022inner}, LLM incorporates diverse environmental feedback to construct inner monologues, thereby formulating effective robot control strategies. 
Furthermore, LP-SLAM \cite{zhang2023lp} proposes a simultaneous localization and mapping (SLAM) system empowered with language perception capabilities, exploiting the potential of ChatGPT. 
PromptCraft \cite{vemprala2023chatgpt}, on the other hand, devises a function library tailored to ChatGPT on the robot platform, streamlining the conversion of user intentions into executable tasks via the underlying backend API.

In addition to directly changing the real environment through interaction with tools in the physical world, 
LLM can also utilize software tools such as search engines \cite{guu2020retrieval, lewis2020retrieval, borgeaud2022improving, sridhar2023hierarchical, furuta2023multimodal, qin2023webcpm, yao2022webshop, nakano2021webgpt, yang2018hotpotqa}, mobile \cite{wang2023enabling, zhang2023mobile}, Microsoft Office \cite{li2023sheetcopilot, zha2023tablegpt}, calculators \cite{chen2023chatcot, parisi2022talm, cobbe2021training}, deep models \cite{patil2023gorilla, yang2023mm, liu2023internchat, ge2023openagi, shen2023hugginggpt, suris2023vipergpt, wu2023visual, gupta2023visual, chen2023language} and other versatile APIs \cite{lu2023chameleon, schick2023toolformer, gou2023critic, liang2023taskmatrix, tang2023toolalpaca, hao2023toolkengpt} to enhance model performance or complete complex workflows through flexible control of the software.
Toolformer \cite{schick2023toolformer} employs a self-supervised methodology to fine-tune the language model, enabling it to acquire the ability to automatically invoke APIs. 
ART \cite{paranjape2023art} leverages CoT \cite{CoT2022} and In-context Learning \cite{chen2023language, mialon2023augmented} techniques to automatically generate multi-step reasoning processes for new tasks, while also selecting and utilizing the most appropriate available tool at each step.
ASH \cite{sridhar2023hierarchical} utilizes LLM for sequence hierarchical decision-making to achieve web navigation tasks. WebGPT \cite{nakano2021webgpt} and WebCPM \cite{qin2023webcpm} use network search to assist in implementing Question Answering tasks. 
In addition, RCI \cite{kim2023language} recursively criticizes and improves itself to execute computer tasks guided by natural language according to the prompting scheme. 
To achieve the analysis and processing of tables, TableGPT \cite{zha2023tablegpt} employs a table encoder to transform tabular data into vector representations, which are then fed into an LLM for inference in combination with user queries.
\subsection{Tool Creation}
The usage of tools is contingent upon the accessibility of external tools. 
Recently, efforts have been made to employ LLM as a tool creator in order to generate tools that can be utilized for diverse requests \cite{cai2023large, lewis2023computegpt, gao2023pal, wang2023voyager, qian2023creator, cai2023low, arora2023language, zhang2023data}.
This development has consequently raised the demands placed on LLM.
And these created tools are typically implemented as Python or SQL functions.
LATM \cite{cai2023large}, for example,  leverages the prowess of GPT-4 to create tools, and the usage of more cost-effective models has shown potential in exhibiting performance on par with larger models for these tool applications.
EVAPORATE \cite{arora2023language} involves the synthesis of multiple functions, which are subsequently utilized at a large scale to efficiently process documents and generate structured views.


\section{Conclusion}
In this paper, we have introduced a structured framework specially designed for LLM-based AI Agents, with an emphasis on their abilities in task planning and tool usage. This framework, coupled with our design of two distinct types of agents assigned for the inference process, allows for a comprehensive evaluation of the capabilities of current open-source LLMs, thereby yielding critical insights into their effectiveness.
Furthermore, our research highlights the significant potential of LLMs in managing complex tasks, revealing the exciting prospects they hold for future research and development. 
As we continue to explore and improve upon these models, we move closer to unlocking their full potential in a wide range of real-world applications.

\section*{Acknowledgements}

This work was conducted collaboratively among the authors.

Hangyu Mao and Rui Zhao led the project, formulating the central idea and laying out the framework for the primary literature review. 

Regarding the literature review phase, the surveys were conducted by various team members. Guoqing Du and Jingqing Ruan explored DNN-based Tool Scheduling by LLMs; Tianpeng Bao and Yihong Chen investigated Physical/Robot Tool Scheduling by LLMs; and Shiwei Shi and Zhiwei Xu handled the survey of API or GUI-based Tool Scheduling by LLMs. Bin Zhang summarized these papers and synthesized an overarching summary.

As for the evaluation phase, Yihong Chen, Tianpeng Bao, Jingqing Ruan, Guoqing Du, Zhiwei Xu, Shiwei Shi, and Bin Zhang performed the experiments and analyzed the data. Hangyu Mao assisted in the analysis of the experimental phenomena and offered constructive suggestions for improvements. Xingyu Zeng and Rui Zhao provided invaluable feedback, contributed to the direction of the research. All authors participated in the discussion.

Regarding the manuscript phase, Hangyu Mao organized the overall chapters of the manuscript and mainly wrote the methodology part, and provided assistance in other parts. Jingqing Ruan and Yihong Chen wrote the evaluation section. Bin Zhang wrote the summary of the literature review. Each author read and approved the final manuscript. 

The authors would like to thank Feng Zhu, Kun Wang, Yuhang Ran, Mengying Xu, Pengfei Jia, and Shaobo Lin for their valuable feedback, discussion, and participation in this project.

\bibliographystyle{IEEEtran}
\bibliography{ref}

@article{zhao2023survey,
  title={A survey of large language models},
  author={Zhao, Wayne Xin and Zhou, Kun and Li, Junyi and Tang, Tianyi and Wang, Xiaolei and Hou, Yupeng and Min, Yingqian and Zhang, Beichen and Zhang, Junjie and Dong, Zican and others},
  journal={arXiv preprint arXiv:2303.18223},
  year={2023}
}

@article{jennings1998roadmap,
  title={A roadmap of agent research and development},
  author={Jennings, Nicholas R and Sycara, Katia and Wooldridge, Michael},
  journal={Autonomous agents and multi-agent systems},
  volume={1},
  pages={7--38},
  year={1998},
  publisher={Springer}
}

@article{jennings1995applying,
  title={Applying agent technology},
  author={Jennings, Nicholas R and Wooldridge, Michael},
  journal={Applied Artificial Intelligence an International Journal},
  volume={9},
  number={4},
  pages={357--369},
  year={1995},
  publisher={Taylor \& Francis}
}

@inproceedings{franklin1996agent,
  title={Is it an Agent, or just a Program?: A Taxonomy for Autonomous Agents},
  author={Franklin, Stan and Graesser, Art},
  booktitle={International workshop on agent theories, architectures, and languages},
  pages={21--35},
  year={1996},
  organization={Springer}
}

@article{castelfranchi1998modelling,
  title={Modelling social action for AI agents},
  author={Castelfranchi, Cristiano},
  journal={Artificial intelligence},
  volume={103},
  number={1-2},
  pages={157--182},
  year={1998},
  publisher={Elsevier}
}

@book{ferber1999multi,
  title={Multi-agent systems: an introduction to distributed artificial intelligence},
  author={Ferber, Jacques and Weiss, Gerhard},
  volume={1},
  year={1999},
  publisher={Addison-wesley Reading}
}

@article{panait2005cooperative,
  title={Cooperative multi-agent learning: The state of the art},
  author={Panait, Liviu and Luke, Sean},
  journal={Autonomous agents and multi-agent systems},
  volume={11},
  pages={387--434},
  year={2005},
  publisher={Springer}
}

@article{pourreza2023din,
  title={Din-sql: Decomposed in-context learning of text-to-sql with self-correction},
  author={Pourreza, Mohammadreza and Rafiei, Davood},
  journal={arXiv preprint arXiv:2304.11015},
  year={2023}
}

@article{gorniak2023vizability,
  title={VizAbility: Multimodal Accessible Data Visualization with Keyboard Navigation and Conversational Interaction},
  author={Gorniak, Joshua and Kim, Yoon and Gwon, Stephen and Wei, Donglai and Kim, Nam Wook},
  journal={arXiv preprint arXiv:2310.09611},
  year={2023}
}

@misc{2023internlm,
    title={InternLM: A Multilingual Language Model with Progressively Enhanced Capabilities},
    author={InternLM Team},
    howpublished = {\url{https://github.com/InternLM/InternLM}},
    year={2023}
}

@article{zeng2022glm,
  title={Glm-130b: An open bilingual pre-trained model},
  author={Zeng, Aohan and Liu, Xiao and Du, Zhengxiao and Wang, Zihan and Lai, Hanyu and Ding, Ming and Yang, Zhuoyi and Xu, Yifan and Zheng, Wendi and Xia, Xiao and others},
  journal={arXiv preprint arXiv:2210.02414},
  year={2022}
}

@article{brown2020language,
  title={Language models are few-shot learners},
  author={Brown, Tom and Mann, Benjamin and Ryder, Nick and Subbiah, Melanie and Kaplan, Jared D and Dhariwal, Prafulla and Neelakantan, Arvind and Shyam, Pranav and Sastry, Girish and Askell, Amanda and others},
  journal={Advances in neural information processing systems},
  volume={33},
  pages={1877--1901},
  year={2020}
}

@article{wei2021finetuned,
  title={Finetuned language models are zero-shot learners},
  author={Wei, Jason and Bosma, Maarten and Zhao, Vincent Y and Guu, Kelvin and Yu, Adams Wei and Lester, Brian and Du, Nan and Dai, Andrew M and Le, Quoc V},
  journal={arXiv preprint arXiv:2109.01652},
  year={2021}
}

@misc{openai2023gpt4,
      title={GPT-4 Technical Report}, 
      author={OpenAI},
      year={2023},
      eprint={2303.08774},
      archivePrefix={arXiv},
      primaryClass={cs.CL}
}

@article{bai2022constitutional,
  title={Constitutional ai: Harmlessness from ai feedback},
  author={Bai, Yuntao and Kadavath, Saurav and Kundu, Sandipan and Askell, Amanda and Kernion, Jackson and Jones, Andy and Chen, Anna and Goldie, Anna and Mirhoseini, Azalia and McKinnon, Cameron and others},
  journal={arXiv preprint arXiv:2212.08073},
  year={2022}
}

@article{ouyang2022training,
  title={Training language models to follow instructions with human feedback},
  author={Ouyang, Long and Wu, Jeffrey and Jiang, Xu and Almeida, Diogo and Wainwright, Carroll and Mishkin, Pamela and Zhang, Chong and Agarwal, Sandhini and Slama, Katarina and Ray, Alex and others},
  journal={Advances in Neural Information Processing Systems},
  volume={35},
  pages={27730--27744},
  year={2022}
}

@article{duan2022survey,
  title={A survey of embodied ai: From simulators to research tasks},
  author={Duan, Jiafei and Yu, Samson and Tan, Hui Li and Zhu, Hongyuan and Tan, Cheston},
  journal={IEEE Transactions on Emerging Topics in Computational Intelligence},
  volume={6},
  number={2},
  pages={230--244},
  year={2022},
  publisher={IEEE}
}

@inproceedings{savva2019habitat,
  title={Habitat: A platform for embodied ai research},
  author={Savva, Manolis and Kadian, Abhishek and Maksymets, Oleksandr and Zhao, Yili and Wijmans, Erik and Jain, Bhavana and Straub, Julian and Liu, Jia and Koltun, Vladlen and Malik, Jitendra and others},
  booktitle={Proceedings of the IEEE/CVF international conference on computer vision},
  pages={9339--9347},
  year={2019}
}

@article{franklin1997autonomous,
  title={Autonomous agents as embodied AI},
  author={Franklin, Stan},
  journal={Cybernetics \& Systems},
  volume={28},
  number={6},
  pages={499--520},
  year={1997},
  publisher={Taylor \& Francis}
}

@article{wake2023chatgpt,
  title={Chatgpt empowered long-step robot control in various environments: A case application},
  author={Wake, Naoki and Kanehira, Atsushi and Sasabuchi, Kazuhiro and Takamatsu, Jun and Ikeuchi, Katsushi},
  journal={arXiv preprint arXiv:2304.03893},
  year={2023}
}

@article{driess2023palm,
  title={Palm-e: An embodied multimodal language model},
  author={Driess, Danny and Xia, Fei and Sajjadi, Mehdi SM and Lynch, Corey and Chowdhery, Aakanksha and Ichter, Brian and Wahid, Ayzaan and Tompson, Jonathan and Vuong, Quan and Yu, Tianhe and others},
  journal={arXiv preprint arXiv:2303.03378},
  year={2023}
}

@inproceedings{liang2023code,
  title={Code as policies: Language model programs for embodied control},
  author={Liang, Jacky and Huang, Wenlong and Xia, Fei and Xu, Peng and Hausman, Karol and Ichter, Brian and Florence, Pete and Zeng, Andy},
  booktitle={2023 IEEE International Conference on Robotics and Automation (ICRA)},
  pages={9493--9500},
  year={2023},
  organization={IEEE}
}

@inproceedings{chen2023open,
  title={Open-vocabulary queryable scene representations for real world planning},
  author={Chen, Boyuan and Xia, Fei and Ichter, Brian and Rao, Kanishka and Gopalakrishnan, Keerthana and Ryoo, Michael S and Stone, Austin and Kappler, Daniel},
  booktitle={2023 IEEE International Conference on Robotics and Automation (ICRA)},
  pages={11509--11522},
  year={2023},
  organization={IEEE}
}

@article{huang2022inner,
  title={Inner monologue: Embodied reasoning through planning with language models},
  author={Huang, Wenlong and Xia, Fei and Xiao, Ted and Chan, Harris and Liang, Jacky and Florence, Pete and Zeng, Andy and Tompson, Jonathan and Mordatch, Igor and Chebotar, Yevgen and others},
  journal={arXiv preprint arXiv:2207.05608},
  year={2022}
}

@inproceedings{brohan2023can,
  title={Do as i can, not as i say: Grounding language in robotic affordances},
  author={Brohan, Anthony and Chebotar, Yevgen and Finn, Chelsea and Hausman, Karol and Herzog, Alexander and Ho, Daniel and Ibarz, Julian and Irpan, Alex and Jang, Eric and Julian, Ryan and others},
  booktitle={Conference on Robot Learning},
  pages={287--318},
  year={2023},
  organization={PMLR}
}

@article{vemprala2023chatgpt,
  title={Chatgpt for robotics: Design principles and model abilities},
  author={Vemprala, Sai and Bonatti, Rogerio and Bucker, Arthur and Kapoor, Ashish},
  journal={Microsoft Auton. Syst. Robot. Res},
  volume={2},
  pages={20},
  year={2023}
}

@article{zhang2023lp,
  title={LP-SLAM: Language-Perceptive RGB-D SLAM system based on Large Language Model},
  author={Zhang, Weiyi and Guo, Yushi and Niu, Liting and Li, Peijun and Zhang, Chun and Wan, Zeyu and Yan, Jiaxiang and Farrukh, Fasih Ud Din and Zhang, Debing},
  journal={arXiv preprint arXiv:2303.10089},
  year={2023}
}

@inproceedings{shah2023lm,
  title={Lm-nav: Robotic navigation with large pre-trained models of language, vision, and action},
  author={Shah, Dhruv and Osi{\'n}ski, B{\l}a{\.z}ej and Levine, Sergey and others},
  booktitle={Conference on Robot Learning},
  pages={492--504},
  year={2023},
  organization={PMLR}
}

@article{brohan2022rt,
  title={Rt-1: Robotics transformer for real-world control at scale},
  author={Brohan, Anthony and Brown, Noah and Carbajal, Justice and Chebotar, Yevgen and Dabis, Joseph and Finn, Chelsea and Gopalakrishnan, Keerthana and Hausman, Karol and Herzog, Alex and Hsu, Jasmine and others},
  journal={arXiv preprint arXiv:2212.06817},
  year={2022}
}

@article{stone2023open,
  title={Open-world object manipulation using pre-trained vision-language models},
  author={Stone, Austin and Xiao, Ted and Lu, Yao and Gopalakrishnan, Keerthana and Lee, Kuang-Huei and Vuong, Quan and Wohlhart, Paul and Zitkovich, Brianna and Xia, Fei and Finn, Chelsea and others},
  journal={arXiv preprint arXiv:2303.00905},
  year={2023}
}

@article{reed2022generalist,
  title={A generalist agent},
  author={Reed, Scott and Zolna, Konrad and Parisotto, Emilio and Colmenarejo, Sergio Gomez and Novikov, Alexander and Barth-Maron, Gabriel and Gimenez, Mai and Sulsky, Yury and Kay, Jackie and Springenberg, Jost Tobias and others},
  journal={arXiv preprint arXiv:2205.06175},
  year={2022}
}

@article{sridhar2023hierarchical,
  title={Hierarchical Prompting Assists Large Language Model on Web Navigation},
  author={Sridhar, Abishek and Lo, Robert and Xu, Frank F and Zhu, Hao and Zhou, Shuyan},
  journal={arXiv preprint arXiv:2305.14257},
  year={2023}
}

@article{furuta2023multimodal,
  title={Multimodal Web Navigation with Instruction-Finetuned Foundation Models},
  author={Furuta, Hiroki and Nachum, Ofir and Lee, Kuang-Huei and Matsuo, Yutaka and Gu, Shixiang Shane and Gur, Izzeddin},
  journal={arXiv preprint arXiv:2305.11854},
  year={2023}
}

@article{qin2023webcpm,
  title={WebCPM: Interactive Web Search for Chinese Long-form Question Answering},
  author={Qin, Yujia and Cai, Zihan and Jin, Dian and Yan, Lan and Liang, Shihao and Zhu, Kunlun and Lin, Yankai and Han, Xu and Ding, Ning and Wang, Huadong and others},
  journal={arXiv preprint arXiv:2305.06849},
  year={2023}
}

@article{yao2022webshop,
  title={Webshop: Towards scalable real-world web interaction with grounded language agents},
  author={Yao, Shunyu and Chen, Howard and Yang, John and Narasimhan, Karthik},
  journal={Advances in Neural Information Processing Systems},
  volume={35},
  pages={20744--20757},
  year={2022}
}

@article{nakano2021webgpt,
  title={Webgpt: Browser-assisted question-answering with human feedback},
  author={Nakano, Reiichiro and Hilton, Jacob and Balaji, Suchir and Wu, Jeff and Ouyang, Long and Kim, Christina and Hesse, Christopher and Jain, Shantanu and Kosaraju, Vineet and Saunders, William and others},
  journal={arXiv preprint arXiv:2112.09332},
  year={2021}
}

@article{zhang2023mobile,
  title={Mobile-Env: A Universal Platform for Training and Evaluation of Mobile Interaction},
  author={Zhang, Danyang and Chen, Lu and Yu, Kai},
  journal={arXiv preprint arXiv:2305.08144},
  year={2023}
}

@inproceedings{wang2023enabling,
  title={Enabling conversational interaction with mobile ui using large language models},
  author={Wang, Bryan and Li, Gang and Li, Yang},
  booktitle={Proceedings of the 2023 CHI Conference on Human Factors in Computing Systems},
  pages={1--17},
  year={2023}
}

@article{kim2023language,
  title={Language models can solve computer tasks},
  author={Kim, Geunwoo and Baldi, Pierre and McAleer, Stephen},
  journal={arXiv preprint arXiv:2303.17491},
  year={2023}
}

@article{li2023sheetcopilot,
  title={SheetCopilot: Bringing Software Productivity to the Next Level through Large Language Models},
  author={Li, Hongxin and Su, Jingran and Chen, Yuntao and Li, Qing and Zhang, Zhaoxiang},
  journal={arXiv preprint arXiv:2305.19308},
  year={2023}
}

@article{rana2023sayplan,
  title={SayPlan: Grounding Large Language Models using 3D Scene Graphs for Scalable Task Planning},
  author={Rana, Krishan and Haviland, Jesse and Garg, Sourav and Abou-Chakra, Jad and Reid, Ian and Suenderhauf, Niko},
  journal={arXiv preprint arXiv:2307.06135},
  year={2023}
}

@article{zha2023tablegpt,
  title={TableGPT: Towards Unifying Tables, Nature Language and Commands into One GPT},
  author={Zha, Liangyu and Zhou, Junlin and Li, Liyao and Wang, Rui and Huang, Qingyi and Yang, Saisai and Yuan, Jing and Su, Changbao and Li, Xiang and Su, Aofeng and others},
  journal={arXiv preprint arXiv:2307.08674},
  year={2023}
}

@article{song2022llm,
  title={Llm-planner: Few-shot grounded planning for embodied agents with large language models},
  author={Song, Chan Hee and Wu, Jiaman and Washington, Clayton and Sadler, Brian M and Chao, Wei-Lun and Su, Yu},
  journal={arXiv preprint arXiv:2212.04088},
  year={2022}
}

@article{chen2023chatcot,
  title={ChatCoT: Tool-Augmented Chain-of-Thought Reasoning on$\backslash$$\backslash$Chat-based Large Language Models},
  author={Chen, Zhipeng and Zhou, Kun and Zhang, Beichen and Gong, Zheng and Zhao, Wayne Xin and Wen, Ji-Rong},
  journal={arXiv preprint arXiv:2305.14323},
  year={2023}
}

@article{bommasani2021opportunities,
  title={On the opportunities and risks of foundation models},
  author={Bommasani, Rishi and Hudson, Drew A and Adeli, Ehsan and Altman, Russ and Arora, Simran and von Arx, Sydney and Bernstein, Michael S and Bohg, Jeannette and Bosselut, Antoine and Brunskill, Emma and others},
  journal={arXiv preprint arXiv:2108.07258},
  year={2021}
}

@article{qin2023tool,
  title={Tool learning with foundation models},
  author={Qin, Yujia and Hu, Shengding and Lin, Yankai and Chen, Weize and Ding, Ning and Cui, Ganqu and Zeng, Zheni and Huang, Yufei and Xiao, Chaojun and Han, Chi and others},
  journal={arXiv preprint arXiv:2304.08354},
  year={2023}
}

@article{mosbach2023few,
  title={Few-shot Fine-tuning vs. In-context Learning: A Fair Comparison and Evaluation},
  author={Mosbach, Marius and Pimentel, Tiago and Ravfogel, Shauli and Klakow, Dietrich and Elazar, Yanai},
  journal={arXiv preprint arXiv:2305.16938},
  year={2023}
}

@article{yang2023harnessing,
  title={Harnessing the power of llms in practice: A survey on chatgpt and beyond},
  author={Yang, Jingfeng and Jin, Hongye and Tang, Ruixiang and Han, Xiaotian and Feng, Qizhang and Jiang, Haoming and Yin, Bing and Hu, Xia},
  journal={arXiv preprint arXiv:2304.13712},
  year={2023}
}

@article{zhang2023one,
  title={One small step for generative ai, one giant leap for agi: A complete survey on chatgpt in aigc era},
  author={Zhang, Chaoning and Zhang, Chenshuang and Li, Chenghao and Qiao, Yu and Zheng, Sheng and Dam, Sumit Kumar and Zhang, Mengchun and Kim, Jung Uk and Kim, Seong Tae and Choi, Jinwoo and others},
  journal={arXiv preprint arXiv:2304.06488},
  year={2023}
}

@article{yu2023nature,
  title={Nature language reasoning, a survey},
  author={Yu, Fei and Zhang, Hongbo and Wang, Benyou},
  journal={arXiv preprint arXiv:2303.14725},
  year={2023}
}

@article{schick2023toolformer,
  title={Toolformer: Language models can teach themselves to use tools},
  author={Schick, Timo and Dwivedi-Yu, Jane and Dess{\`\i}, Roberto and Raileanu, Roberta and Lomeli, Maria and Zettlemoyer, Luke and Cancedda, Nicola and Scialom, Thomas},
  journal={arXiv preprint arXiv:2302.04761},
  year={2023}
}

@article{paranjape2023art,
  title={ART: Automatic multi-step reasoning and tool-use for large language models},
  author={Paranjape, Bhargavi and Lundberg, Scott and Singh, Sameer and Hajishirzi, Hannaneh and Zettlemoyer, Luke and Ribeiro, Marco Tulio},
  journal={arXiv preprint arXiv:2303.09014},
  year={2023}
}

@article{lu2023chameleon,
  title={Chameleon: Plug-and-play compositional reasoning with large language models},
  author={Lu, Pan and Peng, Baolin and Cheng, Hao and Galley, Michel and Chang, Kai-Wei and Wu, Ying Nian and Zhu, Song-Chun and Gao, Jianfeng},
  journal={arXiv preprint arXiv:2304.09842},
  year={2023}
}

@article{gou2023critic,
  title={Critic: Large language models can self-correct with tool-interactive critiquing},
  author={Gou, Zhibin and Shao, Zhihong and Gong, Yeyun and Shen, Yelong and Yang, Yujiu and Duan, Nan and Chen, Weizhu},
  journal={arXiv preprint arXiv:2305.11738},
  year={2023}
}

@article{parisi2022talm,
  title={Talm: Tool augmented language models},
  author={Parisi, Aaron and Zhao, Yao and Fiedel, Noah},
  journal={arXiv preprint arXiv:2205.12255},
  year={2022}
}

@article{cobbe2021training,
  title={Training verifiers to solve math word problems},
  author={Cobbe, Karl and Kosaraju, Vineet and Bavarian, Mohammad and Chen, Mark and Jun, Heewoo and Kaiser, Lukasz and Plappert, Matthias and Tworek, Jerry and Hilton, Jacob and Nakano, Reiichiro and others},
  journal={arXiv preprint arXiv:2110.14168},
  year={2021}
}

@article{tang2023toolalpaca,
  title={ToolAlpaca: Generalized Tool Learning for Language Models with 3000 Simulated Cases},
  author={Tang, Qiaoyu and Deng, Ziliang and Lin, Hongyu and Han, Xianpei and Liang, Qiao and Sun, Le},
  journal={arXiv preprint arXiv:2306.05301},
  year={2023}
}

@article{mialon2023augmented,
  title={Augmented language models: a survey},
  author={Mialon, Gr{\'e}goire and Dess{\`\i}, Roberto and Lomeli, Maria and Nalmpantis, Christoforos and Pasunuru, Ram and Raileanu, Roberta and Rozi{\`e}re, Baptiste and Schick, Timo and Dwivedi-Yu, Jane and Celikyilmaz, Asli and others},
  journal={arXiv preprint arXiv:2302.07842},
  year={2023}
}

@article{patil2023gorilla,
  title={Gorilla: Large language model connected with massive apis},
  author={Patil, Shishir G and Zhang, Tianjun and Wang, Xin and Gonzalez, Joseph E},
  journal={arXiv preprint arXiv:2305.15334},
  year={2023}
}

@article{yang2023mm,
  title={Mm-react: Prompting chatgpt for multimodal reasoning and action},
  author={Yang, Zhengyuan and Li, Linjie and Wang, Jianfeng and Lin, Kevin and Azarnasab, Ehsan and Ahmed, Faisal and Liu, Zicheng and Liu, Ce and Zeng, Michael and Wang, Lijuan},
  journal={arXiv preprint arXiv:2303.11381},
  year={2023}
}

@article{liu2023internchat,
  title={Internchat: Solving vision-centric tasks by interacting with chatbots beyond language},
  author={Liu, Zhaoyang and He, Yinan and Wang, Wenhai and Wang, Weiyun and Wang, Yi and Chen, Shoufa and Zhang, Qinglong and Yang, Yang and Li, Qingyun and Yu, Jiashuo and others},
  journal={arXiv preprint arXiv:2305.05662},
  year={2023}
}

@article{ge2023openagi,
  title={Openagi: When llm meets domain experts},
  author={Ge, Yingqiang and Hua, Wenyue and Ji, Jianchao and Tan, Juntao and Xu, Shuyuan and Zhang, Yongfeng},
  journal={arXiv preprint arXiv:2304.04370},
  year={2023}
}

@article{CoT2022,
  author       = {Jason Wei and
                  Xuezhi Wang and
                  Dale Schuurmans and
                  Maarten Bosma and
                  Brian Ichter and
                  Fei Xia and
                  Ed H. Chi and
                  Quoc V. Le and
                  Denny Zhou},
  title        = {Chain-of-Thought Prompting Elicits Reasoning in Large Language Models},
  journal    = {Neural Information Processing Systems},
  year         = {2022},
  
}

@article{yang2018hotpotqa,
  title={HotpotQA: A dataset for diverse, explainable multi-hop question answering},
  author={Yang, Zhilin and Qi, Peng and Zhang, Saizheng and Bengio, Yoshua and Cohen, William W and Salakhutdinov, Ruslan and Manning, Christopher D},
  journal={arXiv preprint arXiv:1809.09600},
  year={2018}
}

@article{shen2023hugginggpt,
  title={Hugginggpt: Solving ai tasks with chatgpt and its friends in huggingface},
  author={Shen, Yongliang and Song, Kaitao and Tan, Xu and Li, Dongsheng and Lu, Weiming and Zhuang, Yueting},
  journal={arXiv preprint arXiv:2303.17580},
  year={2023}
}

@article{suris2023vipergpt,
  title={Vipergpt: Visual inference via python execution for reasoning},
  author={Sur{\'\i}s, D{\'\i}dac and Menon, Sachit and Vondrick, Carl},
  journal={arXiv preprint arXiv:2303.08128},
  year={2023}
}

@article{wu2023visual,
  title={Visual chatgpt: Talking, drawing and editing with visual foundation models},
  author={Wu, Chenfei and Yin, Shengming and Qi, Weizhen and Wang, Xiaodong and Tang, Zecheng and Duan, Nan},
  journal={arXiv preprint arXiv:2303.04671},
  year={2023}
}

@inproceedings{gupta2023visual,
  title={Visual programming: Compositional visual reasoning without training},
  author={Gupta, Tanmay and Kembhavi, Aniruddha},
  booktitle={Proceedings of the IEEE/CVF Conference on Computer Vision and Pattern Recognition},
  pages={14953--14962},
  year={2023}
}

@inproceedings{chen2023language,
  title={Language Models are Visual Reasoning Coordinators},
  author={Chen, Liangyu and Li, Bo and Shen, Sheng and Yang, Jingkang and Li, Chunyuan and Keutzer, Kurt and Darrell, Trevor and Liu, Ziwei},
  booktitle={ICLR 2023 Workshop on Mathematical and Empirical Understanding of Foundation Models},
  year={2023}
}

@article{wang2023interactive,
  title={Interactive natural language processing},
  author={Wang, Zekun and Zhang, Ge and Yang, Kexin and Shi, Ning and Zhou, Wangchunshu and Hao, Shaochun and Xiong, Guangzheng and Li, Yizhi and Sim, Mong Yuan and Chen, Xiuying and others},
  journal={arXiv preprint arXiv:2305.13246},
  year={2023}
}

@article{hao2023toolkengpt,
  title={ToolkenGPT: Augmenting Frozen Language Models with Massive Tools via Tool Embeddings},
  author={Hao, Shibo and Liu, Tianyang and Wang, Zhen and Hu, Zhiting},
  journal={arXiv preprint arXiv:2305.11554},
  year={2023}
}

@inproceedings{guu2020retrieval,
  title={Retrieval augmented language model pre-training},
  author={Guu, Kelvin and Lee, Kenton and Tung, Zora and Pasupat, Panupong and Chang, Mingwei},
  booktitle={International conference on machine learning},
  pages={3929--3938},
  year={2020},
  organization={PMLR}
}

@article{lewis2020retrieval,
  title={Retrieval-augmented generation for knowledge-intensive nlp tasks},
  author={Lewis, Patrick and Perez, Ethan and Piktus, Aleksandra and Petroni, Fabio and Karpukhin, Vladimir and Goyal, Naman and K{\"u}ttler, Heinrich and Lewis, Mike and Yih, Wen-tau and Rockt{\"a}schel, Tim and others},
  journal={Advances in Neural Information Processing Systems},
  volume={33},
  pages={9459--9474},
  year={2020}
}

@inproceedings{borgeaud2022improving,
  title={Improving language models by retrieving from trillions of tokens},
  author={Borgeaud, Sebastian and Mensch, Arthur and Hoffmann, Jordan and Cai, Trevor and Rutherford, Eliza and Millican, Katie and Van Den Driessche, George Bm and Lespiau, Jean-Baptiste and Damoc, Bogdan and Clark, Aidan and others},
  booktitle={International conference on machine learning},
  pages={2206--2240},
  year={2022},
  organization={PMLR}
}

@article{yu2022survey,
  title={A survey of knowledge-enhanced text generation},
  author={Yu, Wenhao and Zhu, Chenguang and Li, Zaitang and Hu, Zhiting and Wang, Qingyun and Ji, Heng and Jiang, Meng},
  journal={ACM Computing Surveys},
  volume={54},
  number={11s},
  pages={1--38},
  year={2022},
  publisher={ACM New York, NY}
}

@article{hu2021lora,
  title={Lora: Low-rank adaptation of large language models},
  author={Hu, Edward J and Shen, Yelong and Wallis, Phillip and Allen-Zhu, Zeyuan and Li, Yuanzhi and Wang, Shean and Wang, Lu and Chen, Weizhu},
  journal={arXiv preprint arXiv:2106.09685},
  year={2021}
}

@inproceedings{houlsby2019parameter,
  title={Parameter-efficient transfer learning for NLP},
  author={Houlsby, Neil and Giurgiu, Andrei and Jastrzebski, Stanislaw and Morrone, Bruna and De Laroussilhe, Quentin and Gesmundo, Andrea and Attariyan, Mona and Gelly, Sylvain},
  booktitle={International Conference on Machine Learning},
  pages={2790--2799},
  year={2019},
  organization={PMLR}
}

@article{li2021prefix,
  title={Prefix-tuning: Optimizing continuous prompts for generation},
  author={Li, Xiang Lisa and Liang, Percy},
  journal={arXiv preprint arXiv:2101.00190},
  year={2021}
}

@article{liu2021gpt,
  title={GPT understands, too},
  author={Liu, Xiao and Zheng, Yanan and Du, Zhengxiao and Ding, Ming and Qian, Yujie and Yang, Zhilin and Tang, Jie},
  journal={arXiv preprint arXiv:2103.10385},
  year={2021}
}

@article{yao2022react,
  title={React: Synergizing reasoning and acting in language models},
  author={Yao, Shunyu and Zhao, Jeffrey and Yu, Dian and Du, Nan and Shafran, Izhak and Narasimhan, Karthik and Cao, Yuan},
  journal={arXiv preprint arXiv:2210.03629},
  year={2022}
}

@article{khot2022decomposed,
  title={Decomposed prompting: A modular approach for solving complex tasks},
  author={Khot, Tushar and Trivedi, Harsh and Finlayson, Matthew and Fu, Yao and Richardson, Kyle and Clark, Peter and Sabharwal, Ashish},
  journal={arXiv preprint arXiv:2210.02406},
  year={2022}
}

@article{cai2023large,
  title={Large language models as tool makers},
  author={Cai, Tianle and Wang, Xuezhi and Ma, Tengyu and Chen, Xinyun and Zhou, Denny},
  journal={arXiv preprint arXiv:2305.17126},
  year={2023}
}

@article{lewis2023computegpt,
  title={ComputeGPT: A computational chat model for numerical problems},
  author={Lewis, Ryan Hardesty and Jiao, Junfeng},
  journal={arXiv preprint arXiv:2305.06223},
  year={2023}
}

@article{liang2023taskmatrix,
  title={Taskmatrix. ai: Completing tasks by connecting foundation models with millions of apis},
  author={Liang, Yaobo and Wu, Chenfei and Song, Ting and Wu, Wenshan and Xia, Yan and Liu, Yu and Ou, Yang and Lu, Shuai and Ji, Lei and Mao, Shaoguang and others},
  journal={arXiv preprint arXiv:2303.16434},
  year={2023}
}

@inproceedings{gao2023pal,
  title={Pal: Program-aided language models},
  author={Gao, Luyu and Madaan, Aman and Zhou, Shuyan and Alon, Uri and Liu, Pengfei and Yang, Yiming and Callan, Jamie and Neubig, Graham},
  booktitle={International Conference on Machine Learning},
  pages={10764--10799},
  year={2023},
  organization={PMLR}
}

@article{wang2023voyager,
  title={Voyager: An open-ended embodied agent with large language models},
  author={Wang, Guanzhi and Xie, Yuqi and Jiang, Yunfan and Mandlekar, Ajay and Xiao, Chaowei and Zhu, Yuke and Fan, Linxi and Anandkumar, Anima},
  journal={arXiv preprint arXiv:2305.16291},
  year={2023}
}

@article{qian2023creator,
  title={CREATOR: Disentangling Abstract and Concrete Reasonings of Large Language Models through Tool Creation},
  author={Qian, Cheng and Han, Chi and Fung, Yi R and Qin, Yujia and Liu, Zhiyuan and Ji, Heng},
  journal={arXiv preprint arXiv:2305.14318},
  year={2023}
}

@article{cai2023low,
  title={Low-code LLM: Visual Programming over LLMs},
  author={Cai, Yuzhe and Mao, Shaoguang and Wu, Wenshan and Wang, Zehua and Liang, Yaobo and Ge, Tao and Wu, Chenfei and You, Wang and Song, Ting and Xia, Yan and others},
  journal={arXiv preprint arXiv:2304.08103},
  year={2023}
}

@article{arora2023language,
  title={Language Models Enable Simple Systems for Generating Structured Views of Heterogeneous Data Lakes},
  author={Arora, Simran and Yang, Brandon and Eyuboglu, Sabri and Narayan, Avanika and Hojel, Andrew and Trummer, Immanuel and R{\'e}, Christopher},
  journal={arXiv preprint arXiv:2304.09433},
  year={2023}
}

@article{zhang2023data,
  title={Data-Copilot: Bridging Billions of Data and Humans with Autonomous Workflow},
  author={Zhang, Wenqi and Shen, Yongliang and Lu, Weiming and Zhuang, Yueting},
  journal={arXiv preprint arXiv:2306.07209},
  year={2023}
}

@article{touvron2023llama,
  title={Llama: Open and efficient foundation language models},
  author={Touvron, Hugo and Lavril, Thibaut and Izacard, Gautier and Martinet, Xavier and Lachaux, Marie-Anne and Lacroix, Timoth{\'e}e and Rozi{\`e}re, Baptiste and Goyal, Naman and Hambro, Eric and Azhar, Faisal and others},
  journal={arXiv preprint arXiv:2302.13971},
  year={2023}
}

@article{cui2023efficient_alpaca,
  title={Efficient and effective text encoding for chinese llama and alpaca},
  author={Cui, Yiming and Yang, Ziqing and Yao, Xin},
  journal={arXiv preprint arXiv:2304.08177},
  year={2023}
}

@article{xu2023tool,
  title={On the Tool Manipulation Capability of Open-source Large Language Models},
  author={Xu, Qiantong and Hong, Fenglu and Li, Bo and Hu, Changran and Chen, Zhengyu and Zhang, Jian},
  journal={arXiv preprint arXiv:2305.16504},
  year={2023}
}

@article{qin2023toolllm,
  title={ToolLLM: Facilitating Large Language Models to Master 16000+ Real-world APIs},
  author={Qin, Yujia and Liang, Shihao and Ye, Yining and Zhu, Kunlun and Yan, Lan and Lu, Yaxi and Lin, Yankai and Cong, Xin and Tang, Xiangru and Qian, Bill and others},
  journal={arXiv preprint arXiv:2307.16789},
  year={2023}
}

@article{li2023api,
  title={Api-bank: A benchmark for tool-augmented llms},
  author={Li, Minghao and Song, Feifan and Yu, Bowen and Yu, Haiyang and Li, Zhoujun and Huang, Fei and Li, Yongbin},
  journal={arXiv preprint arXiv:2304.08244},
  year={2023}
}

\appendix
\section{Technical Appendices and Supplementary Material}

\subsection{Detailed Dataset Description}
\label{app:eva_datasets}



 \textbf{Simple SQL queries}: These queries typically involve basic operations such as SELECT, FROM, WHERE, GROUP BY, etc. They are used to retrieve, filter, group, and sort data from a single table. We give the Schema of two tables in the SQL database in Table~\ref{tab:schema_simple1} and~\ref{tab:schema_simple2}.

\begin{table}[h!]
\centering
\begin{minipage}{.5\linewidth}
\centering
\caption{Schema of the Person table}
\begin{tabular}{ll}
\toprule
\multicolumn{2}{c}{\textbf{Person}} \\
\midrule
\textbf{Column Name} & \textbf{Type} \\
\midrule
id & TEXT \\
name & TEXT \\
age & INTEGER \\
sex & TEXT \\
school & TEXT \\
phone & TEXT \\
qualifications & TEXT \\
ability & TEXT \\
\bottomrule
\end{tabular}
\label{tab:schema_simple1}
\end{minipage}%
\begin{minipage}{.5\linewidth}
\centering
\caption{Schema of the School table}
\begin{tabular}{ll}
\toprule
\multicolumn{2}{c}{\textbf{School}} \\
\midrule
\textbf{Column Name} & \textbf{Type} \\
\midrule
id & TEXT \\
name & TEXT \\
info\_985 & TEXT \\
info\_211 & TEXT \\
\bottomrule
\end{tabular}
\label{tab:schema_simple2}
\end{minipage}
\end{table}


 \textbf{Complex nested SQL queries}: These queries contain subqueries, which are SQL queries nested inside a larger query. Nested queries can be used in various clauses such as SELECT, FROM, WHERE, and HAVING. They provide a way to perform multiple operations or calculations across multiple tables.
We give the Schema of two tables in the SQL database in Table~\ref{tab:schema_nest1},~\ref{tab:schema_nest2},~\ref{tab:schema_nest3}, and~\ref{tab:schema_nest4}.
\begin{table}[h!]
\centering
\begin{minipage}{.5\linewidth}
\centering
\caption{Schema of GoldenMelodyAwards}
\begin{tabular}{ll}
\toprule
\multicolumn{2}{c}{\textbf{GoldenMelodyAwards}} \\
\midrule
\textbf{Column Name} & \textbf{Type} \\
\midrule
Nominated\_Count & INTEGER \\
Competing\_Count & INTEGER \\
Awards\_Count & INTEGER \\
Award\_Name & TEXT \\
Host & TEXT \\
Year & TIME \\
\bottomrule
\end{tabular}
\label{tab:schema_nest1}
\end{minipage}%
\begin{minipage}{.5\linewidth}
\centering
\caption{Schema of the AwardNominees table}
\begin{tabular}{ll}
\toprule
\multicolumn{2}{c}{\textbf{AwardNominees}} \\
\midrule
\textbf{Column Name} & \textbf{Type} \\
\midrule
Singer\_ID & INTEGER \\
Nominated\_Work & TEXT \\
Award\_Name & TEXT \\
Award\_Edition\_ID & INTEGER \\
\bottomrule
\end{tabular}
\label{tab:schema_nest2}
\end{minipage}
\end{table}

\begin{table}[h!]
\centering
\begin{minipage}{.5\linewidth}
\centering
\caption{Schema of the Singers table}
\begin{tabular}{ll}
\toprule
\multicolumn{2}{c}{\textbf{Singers}} \\
\midrule
\textbf{Column Name} & \textbf{Type} \\
\midrule
Name & TEXT \\
Song\_Count & INTEGER \\
Album\_Count & INTEGER \\
Fan\_Count & INTEGER \\
Gender & TEXT \\
Singer\_ID & INTEGER \\
\bottomrule
\end{tabular}
\label{tab:schema_nest3}
\end{minipage}%
\begin{minipage}{.5\linewidth}
\centering
\caption{Schema of the RecordCompanies table}
\begin{tabular}{ll}
\toprule
\multicolumn{2}{c}{\textbf{RecordCompanies}} \\
\midrule
\textbf{Column Name} & \textbf{Type} \\
\midrule
Record\_Company & TEXT \\
Signing\_Date & TIME \\
Singer\_ID & INTEGER \\
\bottomrule
\end{tabular}
\label{tab:schema_nest4}
\end{minipage}
\end{table}


 \textbf{Complex nested queries utilizing multiple tools}: These are advanced queries that involve multiple tools, such as SQL queries, python code generation, user-defined functions, etc. We give the Schema of two tables in the SQL database in Table~\ref{tab:schema_nest_multi1}, and~\ref{tab:schema_nest_multi2}.
 For verifying the planning ability of the LLM-based AI agents, we select this type of query.


\begin{table}[h!]
\centering
\begin{minipage}{.5\linewidth}
\centering
\caption{Schema of the Journal table}
\begin{tabular}{ll}
\toprule
\multicolumn{2}{c}{\textbf{Journal}} \\
\midrule
\textbf{Column Name} & \textbf{Type} \\
\midrule
Name  & TEXT \\
First\_Issue\_Date & TIME \\
Journal\_ID & INTEGER \\
Category  & TEXT \\
Sponsor\_Organization & TEXT \\
Country  & TEXT \\ s
Language  & TEXT \\
Publication\_Count & INTEGER \\
\bottomrule
\end{tabular}
\label{tab:schema_nest_multi1}
\end{minipage}%
\begin{minipage}{.5\linewidth}
\centering
\caption{Schema of the CoverPersonality table}
\begin{tabular}{ll}
\toprule
\multicolumn{2}{c}{\textbf{CoverPersonality}} \\
\midrule
\textbf{Column Name} & \textbf{Type} \\
\midrule
Person\_ID & INTEGER \\
Journal\_ID & INTEGER \\
Count & INTEGER \\
\bottomrule
\end{tabular}
\label{tab:schema_nest_multi2}
\end{minipage}
\end{table}

\clearpage
\subsection{Prompts Design}
\label{app:prompts}


\begin{figure}[h!]
\centering
\caption{The evaluation prompt for tool order planning.}
\includegraphics[width=1.0\columnwidth]{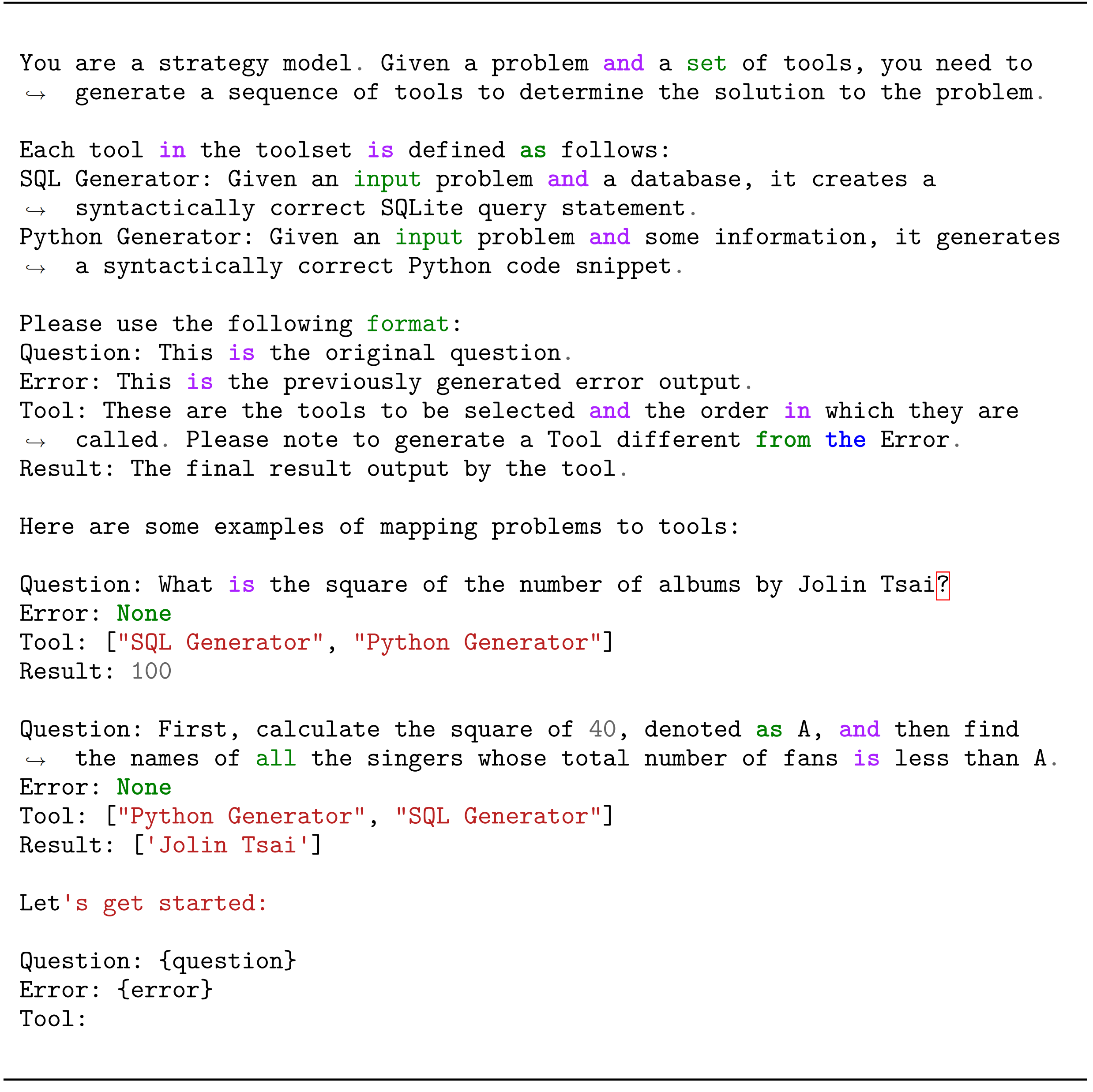}
\label{tab:prompt_plan_tool_name}
\end{figure}


\begin{figure}[t]
\centering
    \caption{The evaluation prompt for tool order and subtask description planning.}
    \includegraphics[width=1.0\columnwidth]{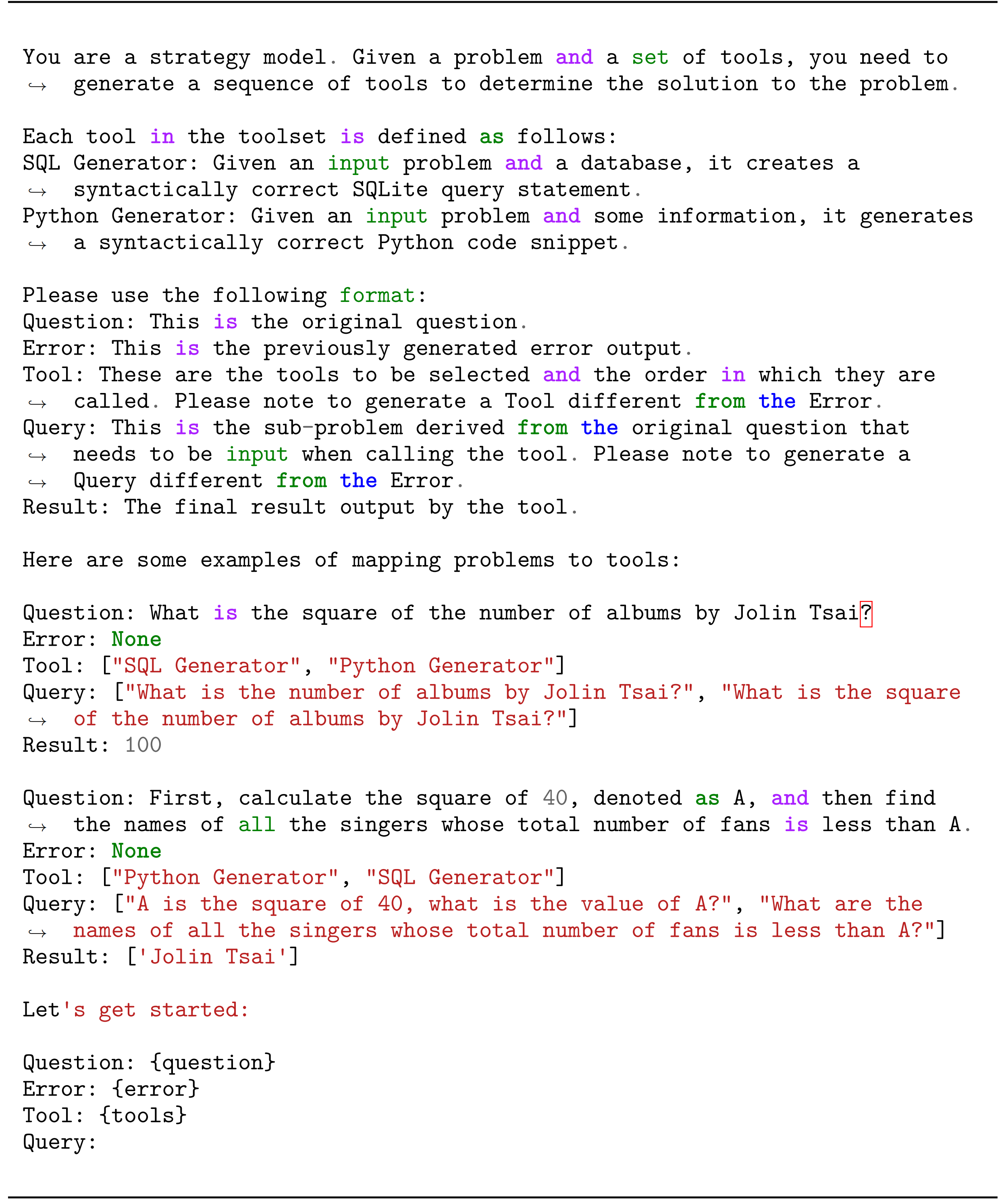}
    \label{tab:prompt_plan_tool_name_query}
\end{figure}


\begin{figure}[t]
\centering
    \caption{The evaluation prompt for one-step tool-subtask pair planning.}
    \includegraphics[width=1.0\columnwidth]{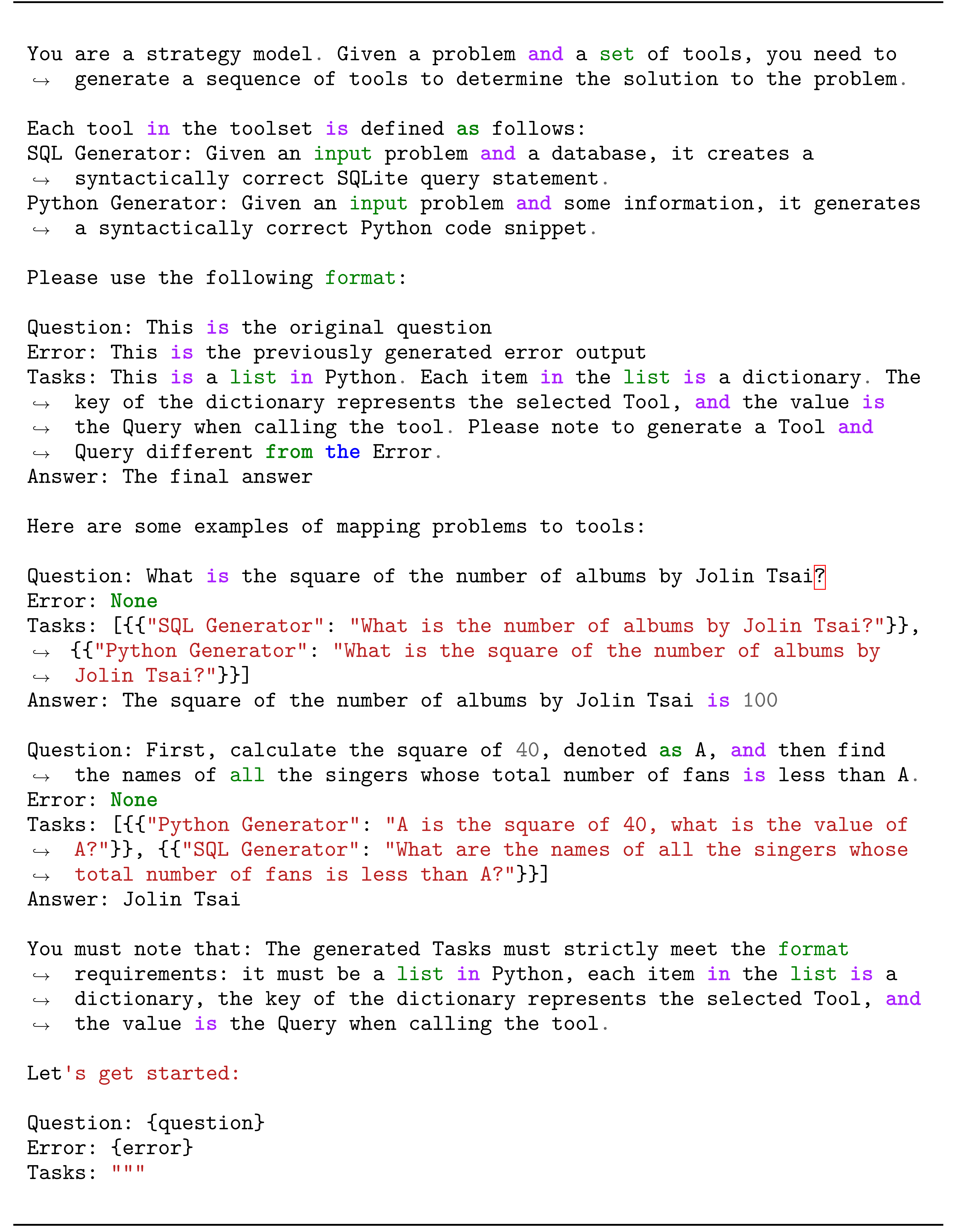}
    \label{tab:prompt_plan_oneshot}
\end{figure}


\begin{figure}[t]
\centering
    \caption{The prompt added to Figure~\ref{tab:prompt_plan_oneshot} for tool-subtask pair planning with other unrelated tools.}
    \includegraphics[width=1.0\columnwidth]{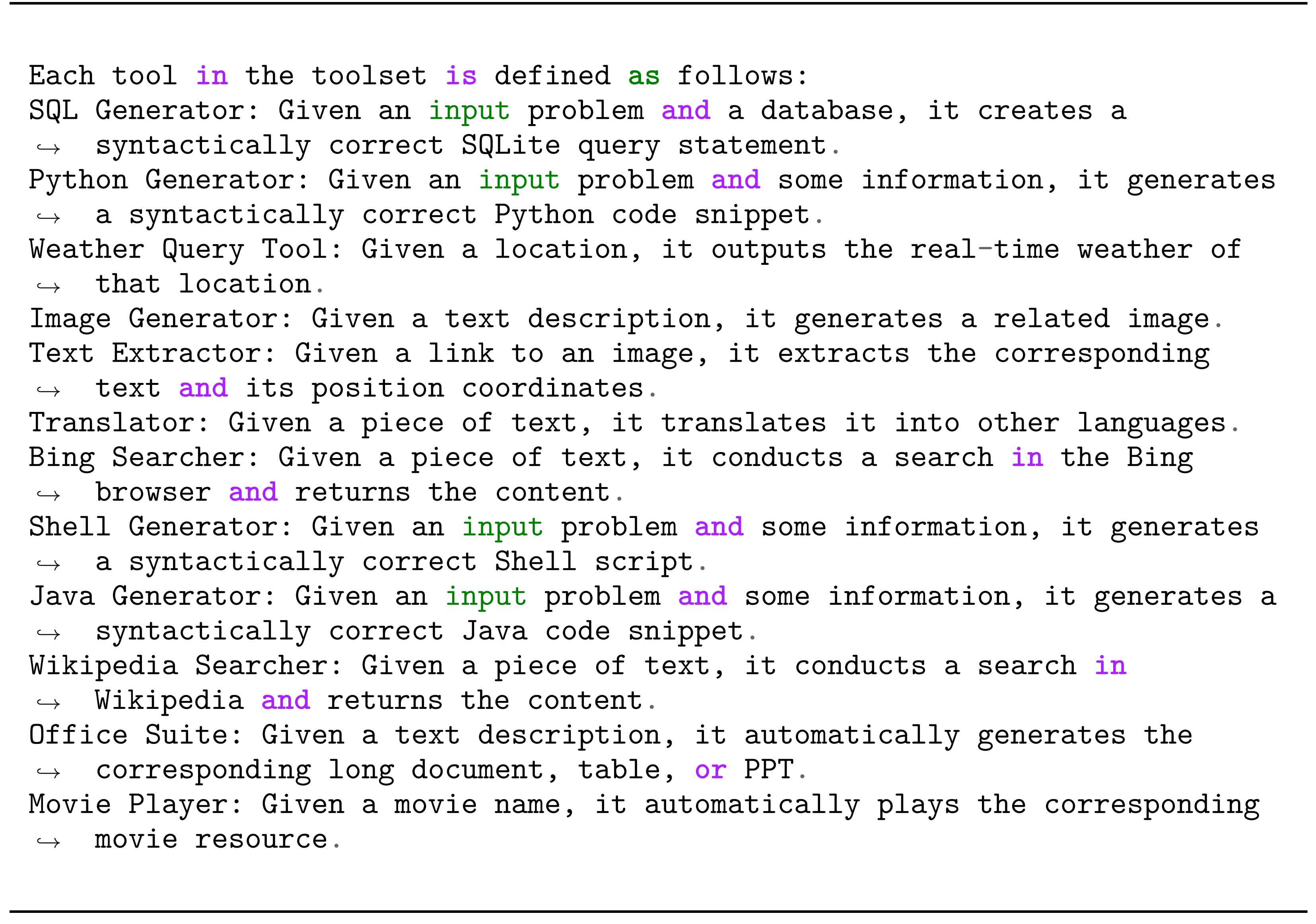}
    \label{tab:prompt_plan_others}
\end{figure}


\begin{figure}[t]
\centering
    \caption{The prompt for the tool-subtask pair generation with~\nameSeq.}
    \includegraphics[width=0.96\columnwidth]{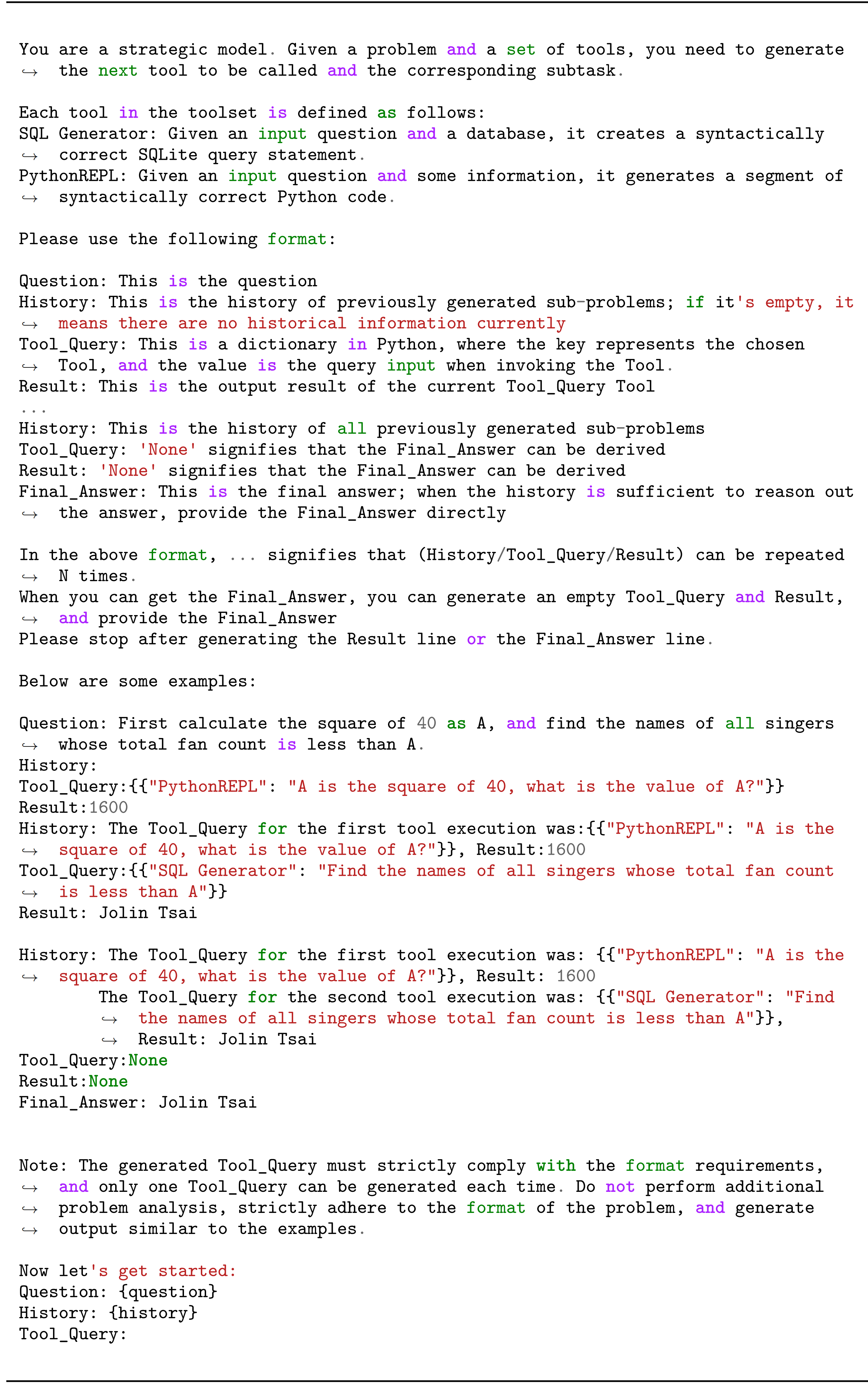}
    \label{tab:prompt_sa_planning}
\end{figure}



\begin{figure}[t]
\centering
    \caption{The evaluation prompt for simple SQL questions.}
    \includegraphics[width=1.0\columnwidth]{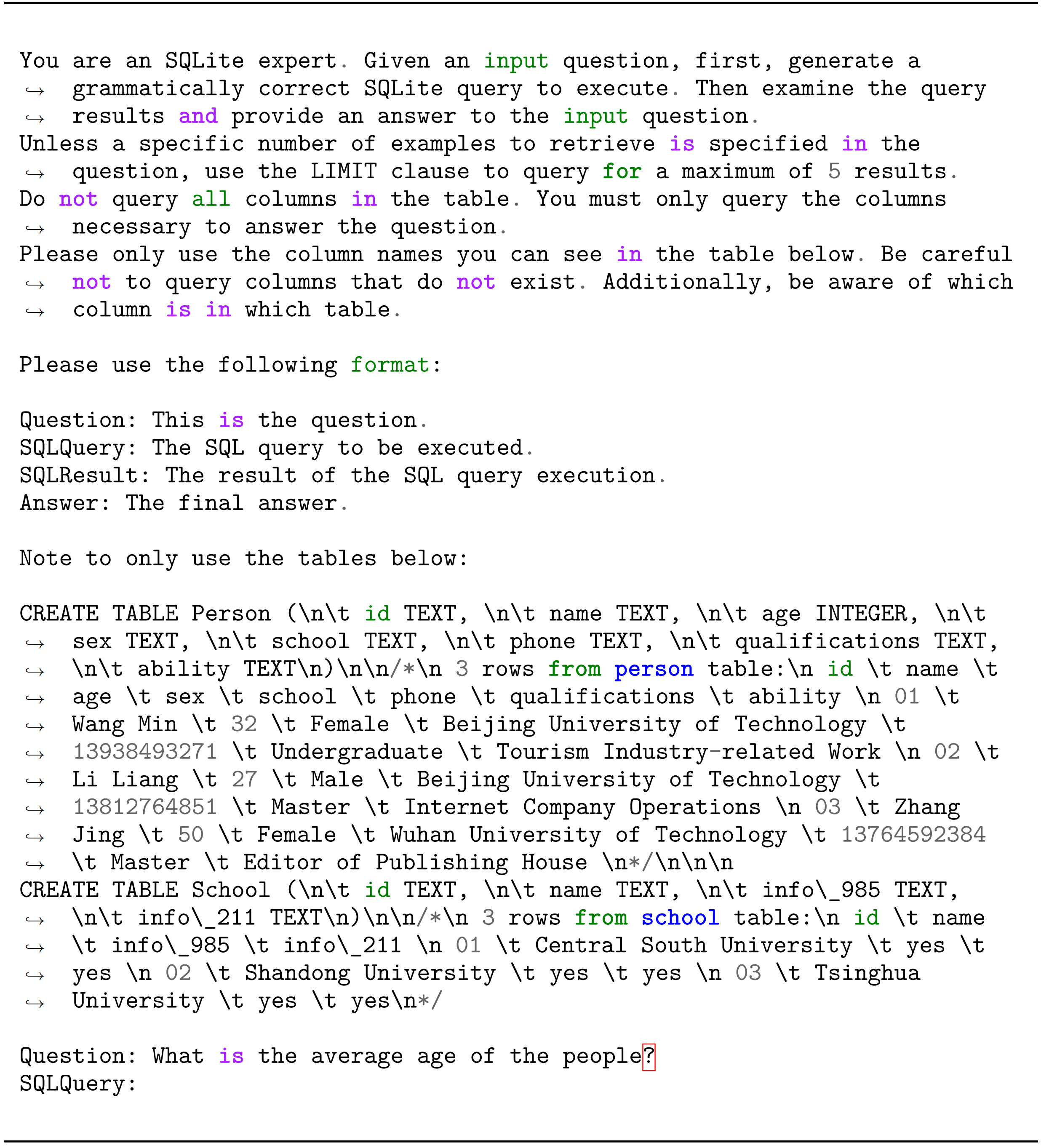}
    \label{tab:prompt_simple_sql}
\end{figure}



\begin{figure}[t]
\centering
    \caption{The evaluation prompt for complex nested SQL questions.}
    \includegraphics[width=1.0\columnwidth]{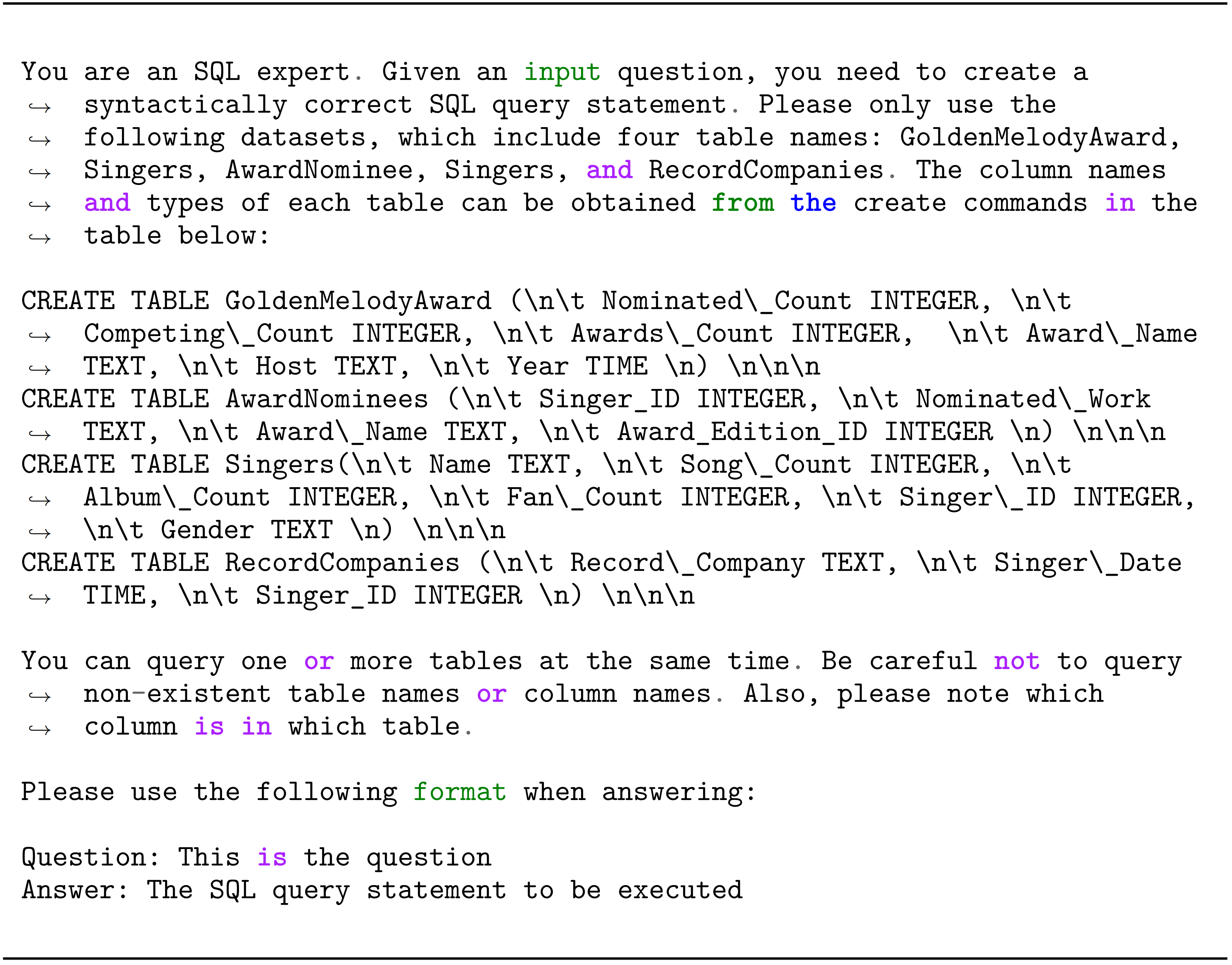}
    \label{tab:prompt_nested_sql}
\end{figure}


\begin{figure}[t]
\centering
    \caption{The evaluation CoT-based prompt for complex nested SQL questions.}
    \includegraphics[width=1.0\columnwidth]{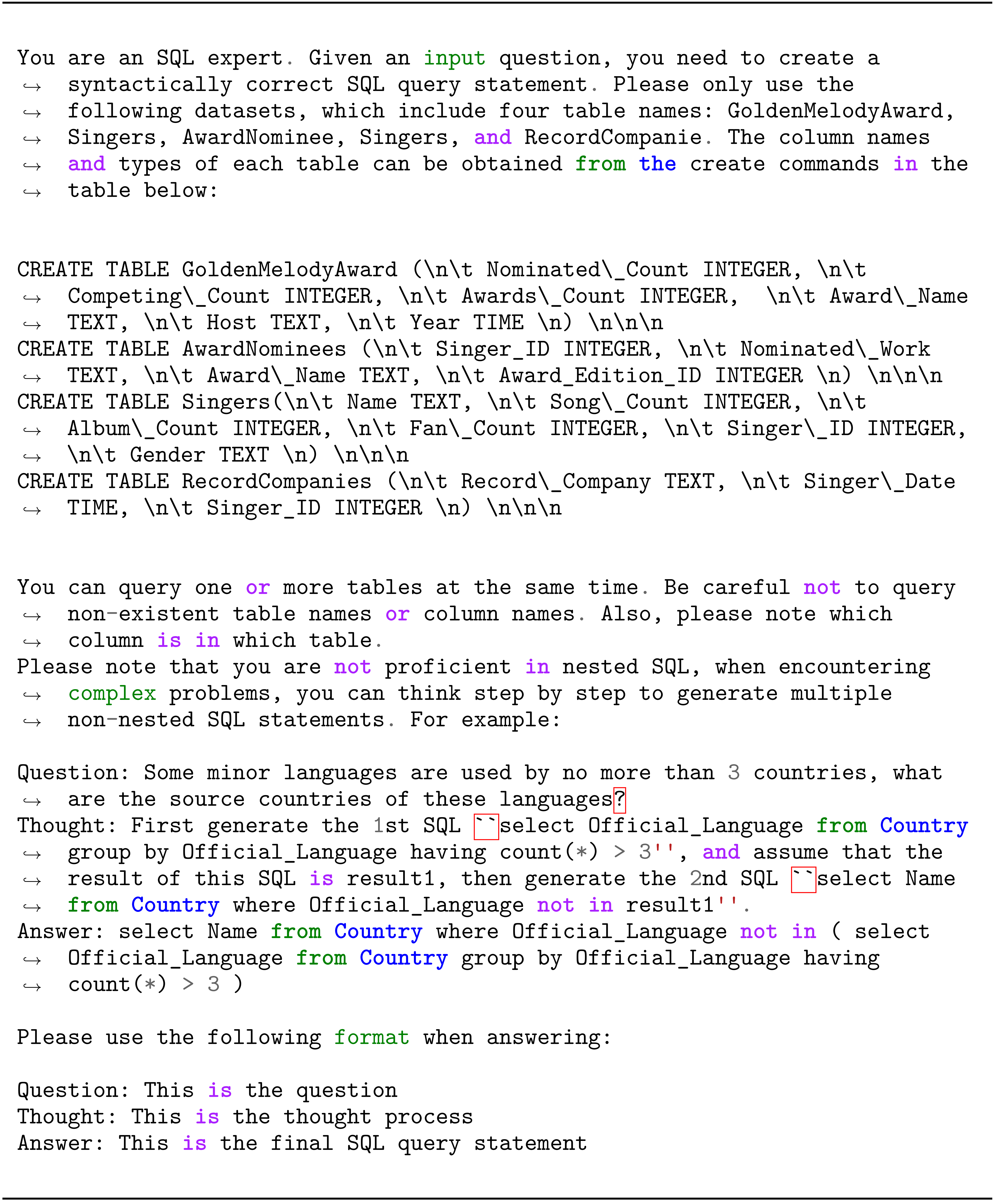}
    \label{tab:prompt_nested_sql_cot}
\end{figure}


\begin{figure}[t]
\centering
    \caption{The evaluation prompt for mathematical  questions.}
    \includegraphics[width=1.0\columnwidth]{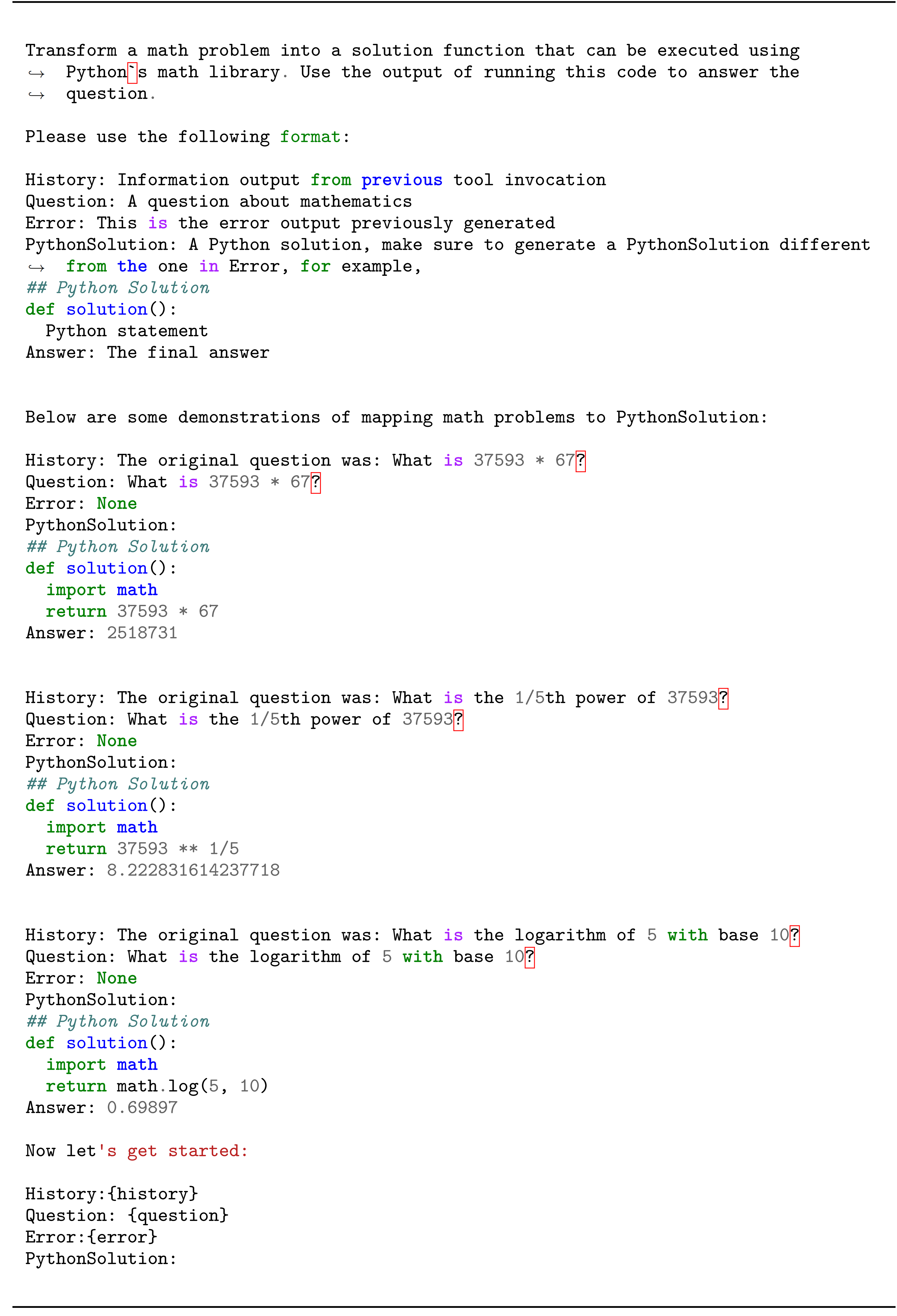}
    \label{tab:prompt_math_qa}
\end{figure}


\begin{figure}[t]
\centering
    \caption{The system prompt for one-step agent.}
    \includegraphics[width=1.0\columnwidth]{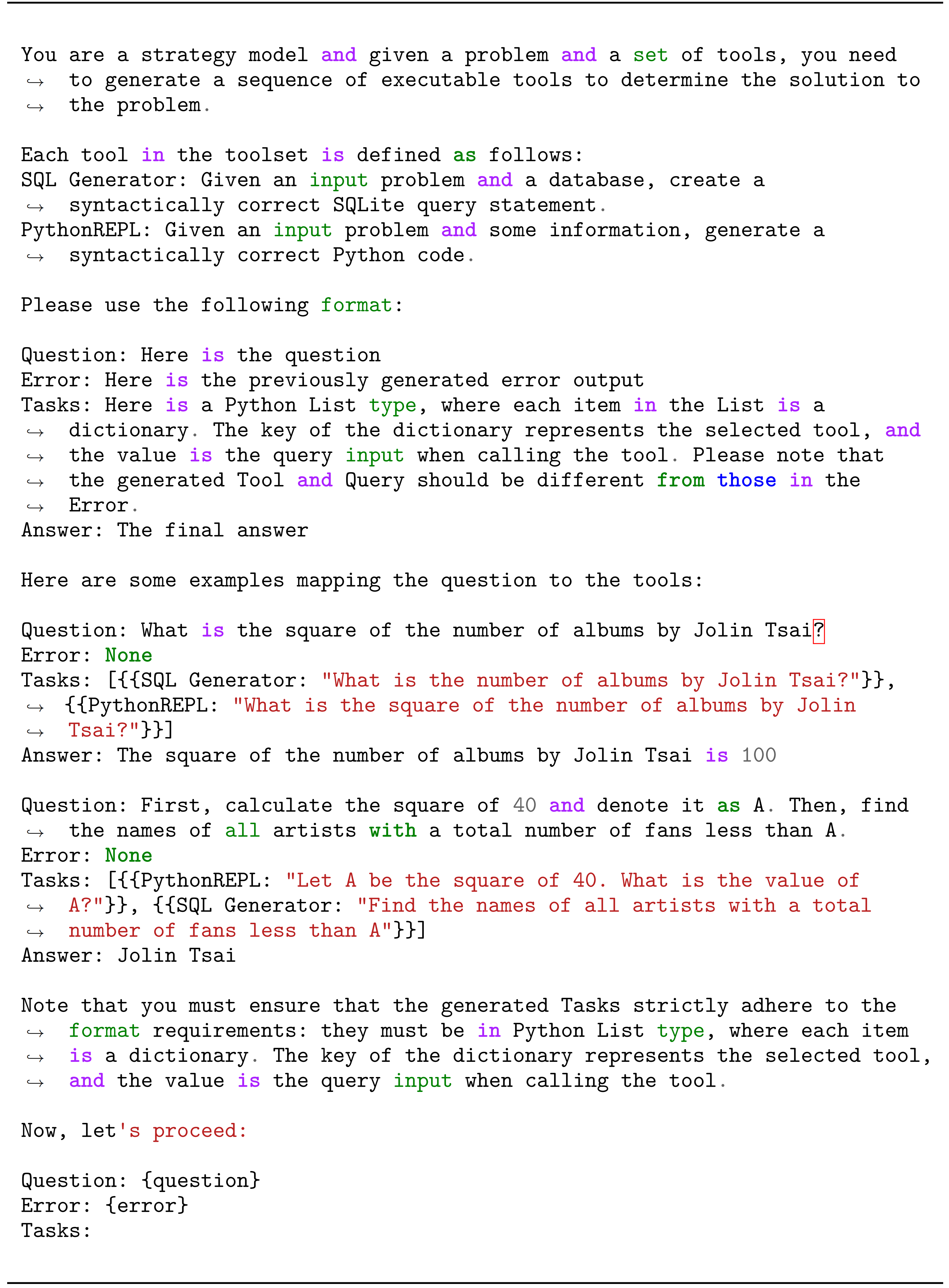}
    \label{tab:prompt_end2end_onestep}
\end{figure}


\begin{figure}[t]
\centering
    \caption{The system prompt for the sequential agent.}
    \includegraphics[width=1.0\columnwidth]{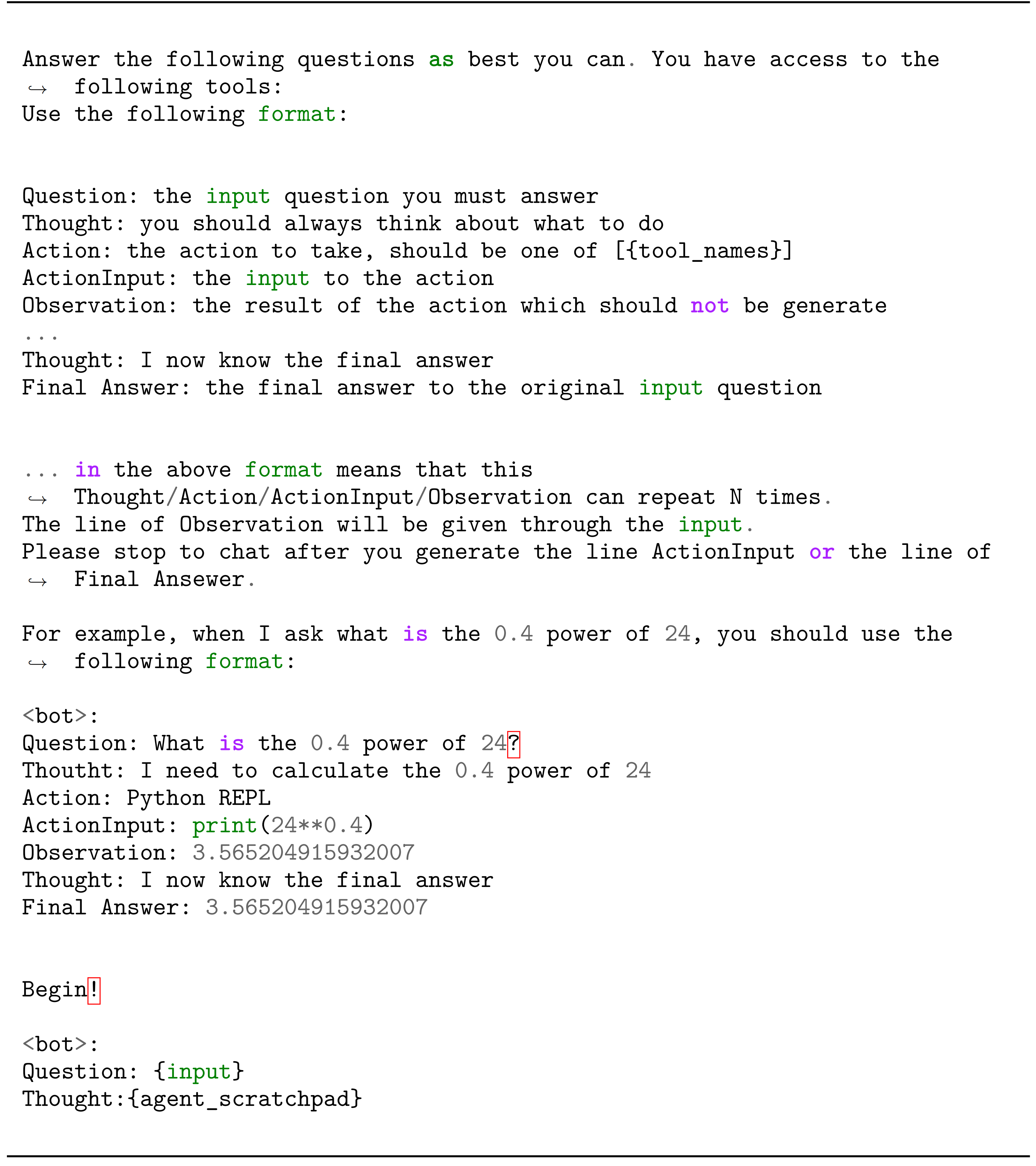}
    \label{tab:prompt_end2end_sequential}
\end{figure}


\end{document}